\documentclass[11pt]{article}

\usepackage[preprint]{acl}

\usepackage{times}
\usepackage{latexsym}

\usepackage[T1]{fontenc}

\usepackage[utf8]{inputenc}

\usepackage{microtype}

\usepackage{inconsolata}

\usepackage{graphicx}

\usepackage{hyperref}
\usepackage{url}

\usepackage{algorithm}
\usepackage{algpseudocode}
\usepackage{subcaption}

\usepackage{multirow}
\usepackage{booktabs} 
\usepackage{amsmath} 
\usepackage{adjustbox}

\title{JudgeAgent: Beyond Static Benchmarks for Knowledge-Driven and Dynamic LLM Evaluation}

\author{
 \textbf{Zhichao Shi\textsuperscript{1,2,3,4}}\thanks{Both authors contributed equally to this research.},
 \textbf{Xuhui Jiang\textsuperscript{1,2}}\footnotemark[1],
 \textbf{Chengjin Xu\textsuperscript{1,2}},
 \textbf{Cangli Yao\textsuperscript{1,2}},
\\
 \textbf{Shengjia Ma\textsuperscript{1,2}},
 \textbf{Yinghan Shen\textsuperscript{4}},
 \textbf{Zixuan Li\textsuperscript{4}},
 \textbf{Jian Guo\textsuperscript{2}},
 \textbf{Yuanzhuo Wang\textsuperscript{4}}\thanks{Corresponding author},
\\
 \textsuperscript{1}DataArc Tech Ltd.\\
 \textsuperscript{2}IDEA Research, International Digital Economy Academy\\
 \textsuperscript{3}School of Advanced Interdisciplinary Sciences, UCAS\\
 \textsuperscript{4}State Key Lab of AI Safety, Institute of Computing Technology, CAS
}

\begin{document}
\maketitle
\begin{abstract}

Current evaluation methods for large language models (LLMs) primarily rely on static benchmarks, presenting two major challenges: limited knowledge coverage and fixed difficulties that mismatch with the evaluated LLMs.
These limitations lead to superficial assessments of LLM knowledge, thereby impeding the targeted model optimizations.
To bridge this gap, we propose JudgeAgent, a knowledge-driven and dynamic evaluation framework for LLMs.
To address the challenge of limited knowledge coverage, JudgeAgent leverages LLM agents equipped with context graphs to traverse knowledge structures systematically for question generation.
Furthermore, to mitigate data contamination and difficulty mismatch, it adopts a difficulty-adaptive and multi-turn interview mechanism.
Thereby, JudgeAgent can achieve comprehensive evaluations and facilitate more effective improvement of LLMs.
Empirical results demonstrate that JudgeAgent enables more comprehensive evaluations and facilitates effective model iterations, highlighting the potential of this knowledge-driven and dynamic evaluation paradigm.
The source code is available on \url{https://github.com/DataArcTech/JudgeAgent}. 
\end{abstract}

\section{Introduction}
\label{sec: intro}


Evaluating Large Language Models (LLMs) to verify their knowledge is a crucial step for their successful application across various domains.\citep{tang2024llms, yuan2023advanced, shi2024drug, wang2025evaluating}. 
Current evaluation methods mainly rely on static benchmarks\citep{clark2018think, hendrycksmeasuring, huang2023c, lin2022truthfulqa, cobbe2021training, multihoprag}, where LLMs are evaluated by their performance on answering predefined static questions.
In the early stages, benefiting from the controllable question quality and rapid workflows of static benchmarking, developers can quickly grasp the core competencies of evolving LLMs, thereby accelerating iteration cycles.
However, due to the static nature\citep{survey_llm_judge, ko2024hierarchical}, these benchmarks have become saturated more and more quickly in recent years. 
For example, it took 3 years for MMLU\citep{mmlu} to reach 80\% accuracy record, while GPQA\citep{gpqa} achieved the same record just within one year.
The static nature restricts the evaluations within a predefined and limited knowledge scope\citep{wangmint, kwan2024mt}, and increases the risk of data contamination\citep{schaeffer2023pretraining, oren2023proving}, resulting in LLMs being misapplied in scenarios out of their knowledge.

\begin{figure*}
    \centering
    \includegraphics[width=0.9\linewidth]{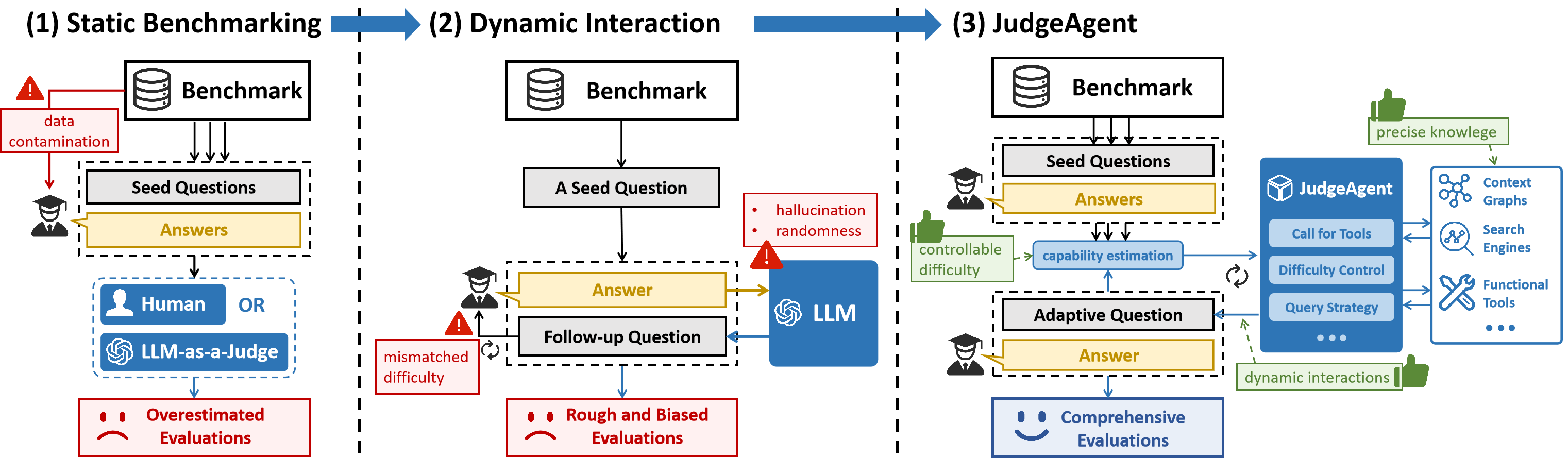}
    \caption{The difference between JudgeAgent and current evaluation paradigms.}
    \label{fig: paradigm}
    \vspace{-1.5em}
\end{figure*}

Consequently, researchers attempted to modify static benchmarks using LLMs\citep{bai2023benchmarking, safetyquizzer}. 
However, they still face two major challenges.
\textbf{(1) Limited knowledge coverage}: Constrained by the evaluator LLM's knowledge limitations and inherent randomness, the knowledge extended by the LLM from static benchmarks remains uncontrollable and limited, making it difficult to achieve comprehensive knowledge evaluations.
\textbf{(2) Mismatch between question difficulty and the evaluated LLMs}: Current mainstream methods lack adaptive adjustments to question difficulty. 
Thus, the question difficulty may significantly deviate from the actual knowledge mastery of the evaluated LLMs, making the evaluations harder to precisely reflect the model's capabilities.
Thus, a new knowledge-driven and dynamic evaluation framework is needed.

In this paper, we propose JudgeAgent, a knowledge-driven and dynamic evaluation framework for LLMs. 
To address the challenge of limited knowledge coverage, JudgeAgent leverages LLM agents with context graphs to traverse knowledge structures at greater breadth and depth.
Through graph sampling strategies and knowledge-driven data synthesis, JudgeAgent generates questions with broader and deeper knowledge coverage, further exploring potential knowledge deficiencies in evaluated LLMs.
Furthermore, to mitigate data contamination and difficulty mismatch, JudgeAgent dynamically adjusts the capability estimation and generates difficulty-adaptive questions based on the target's responses in multi-turn interviews.
Thereby, JudgeAgent can provide comprehensive evaluations and facilitate more effective subsequent improvement of evaluated LLMs.

Extensive experiments and analysis validate that JudgeAgent can provide effective and precise evaluation results. 
Our in-depth analysis also reveals the significant potential of this evaluation paradigm.
It offers valuable insights into how additional knowledge-driven and dynamic strategies for evaluator LLMs can be designed, which enhance the quality of dynamic evaluation results.

In summary, our contributions are as follows:
\begin{itemize}
    \item We introduce JudgeAgent, a knowledge-driven and dynamic evaluation framework for LLMs. 
    JudgeAgent leverages LLM agents with context graphs to traverse knowledge structures and perform dynamic multi-turn interviews, achieving comprehensive knowledge evaluations.
    \item We propose a difficulty-adaptive and multi-turn evaluation mechanism that simulates an interactive interview, enabling the evaluations to more precisely reflect the knowledge mastery of the evaluated LLM.
    \item We conduct extensive experiments to validate the effectiveness of JudgeAgent and highlight its significant potential for the dynamic evaluation paradigm.
\end{itemize}
\section{Related Works}
\label{sec: related_works}

\subsection{Static Benchmark-based Evaluation}
\label{subsec: 2_1_static}
These methods employ pre-constructed benchmarks to evaluate LLMs, using formats such as multiple-choice, question-answer(Q\&A), or prompts for performing tasks. For multiple-choice \citep{clark2018think, hendrycksmeasuring, huang2023c} or Q\&A \citep{lin2022truthfulqa, cobbe2021training, multihoprag}, the benchmark provides correct answers and assesses the target by the accuracy.
For task-execution format, the benchmark measures the model by the metrics assessed by human \citep{chang2024survey} or LLM-as-a-judge \citep{pandalm, mtbench}. 
These methods ensure controllable question quality, but they are also susceptible to data contamination, which undermines the validity of evaluations.

\subsection{Dynamic LLM-based Evaluation}
\label{subsec: 2_2_dynamic}

\begin{figure*}
    \centering
    \includegraphics[width=0.8\linewidth]{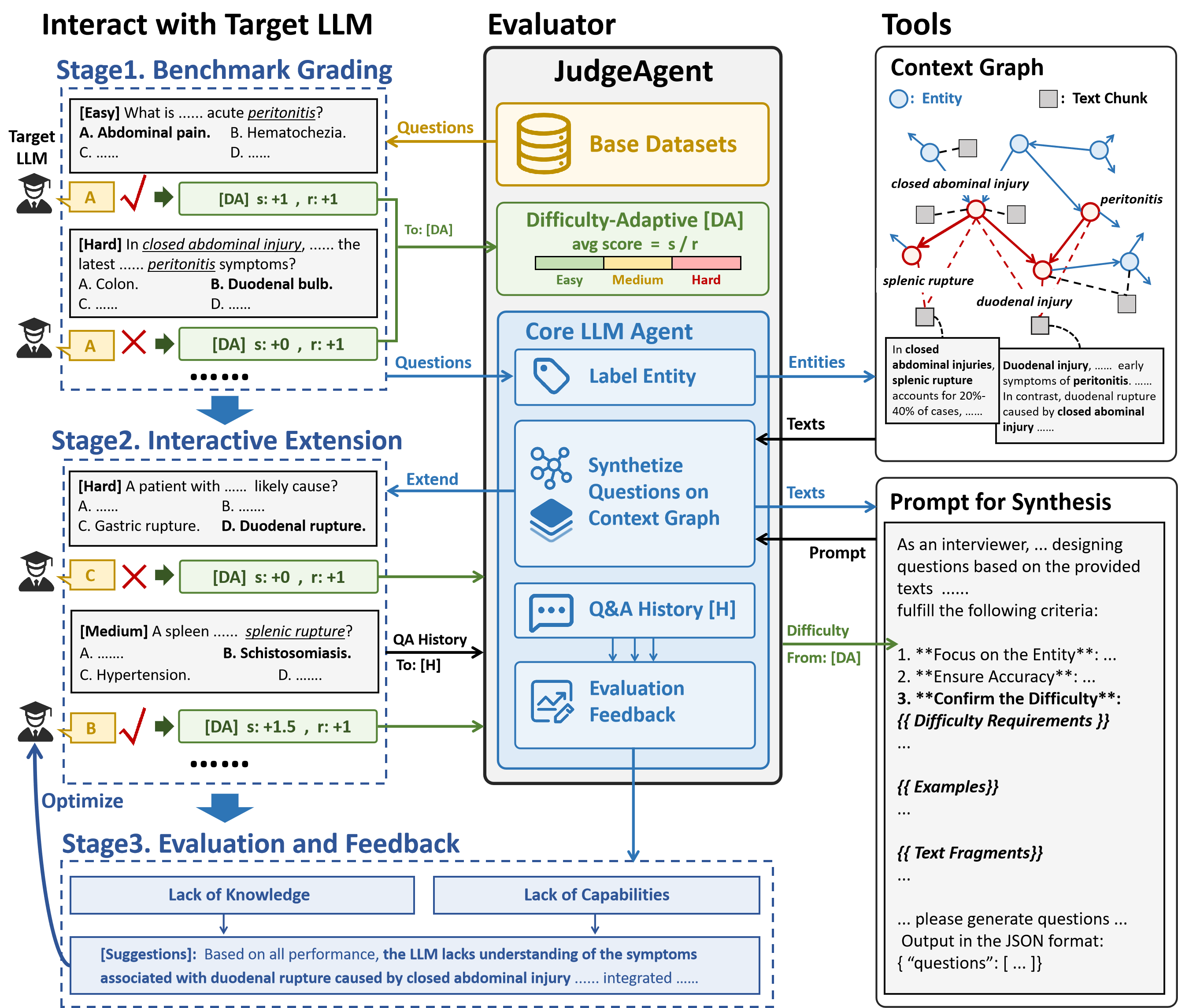}
    \caption{The framework of JudgeAgent. The left part is the interaction process. The central part is the composition of JudgeAgent. The right part presents the tools of JudgeAgent.}
    \label{fig: framework}
    \vspace{-1em}
\end{figure*}


With the deepening application of LLMs, in addition to mainstream static methods, recent researchers have begun to leverage LLMs to dynamically generate evaluation questions.
\citet{bai2023benchmarking} employs LLMs as examiners to generate questions based on Google Trends categories. 
SafetyQuizzer\citep{safetyquizzer} formulates questions with current events from search engines to maximize the timeliness. 
KIEval\citep{kieval}, and LLM-as-an-Interviewer \citep{kim-etal-2025-llm-interviewer} leverage LLMs to generate follow-up questions based on the target's responses in multi-turn interactions.

Compared to static methods, these dynamic methods address the challenge of data contamination and provide more genuine evaluations.
However, the generations of these methods rely mainly on the LLM, resulting in limited knowledge coverage and pool quality control.
Moreover, these methods lack adaptive adjustments of question difficulty, resulting in a deviation between the evaluations and the actual capabilities of target LLMs. 
This mismatch leads to biased subsequent optimization efforts and reduced efficiency.

Thus, we propose a knowledge-driven and dynamic evaluation framework named JudgeAgent.
To expand knowledge coverage, JudgeAgent leverages LLM agents with context graphs to traverse knowledge structures and generate questions.
To mitigate data contamination and difficulty mismatch, JudgeAgent adopts a difficulty-adaptive multi-turn interview mechanism to conduct dynamic evaluations.
Thus, JudgeAgent provides comprehensive evaluations and facilitates subsequent improvement of LLMs.
\section{Methodology}
\label{sec: method}

We introduce JudgeAgent, a knowledge-driven dynamic evaluation framework, which utilizes LLM agents with context graphs to simulate the entire interview of target LLMs.
The workflow comprises three core components: 
\textbf{(1) Benchmark Grading}: Fundamentally assessing target LLMs through testing on public static benchmarks. 
\textbf{(2) Interactive Extension}: Using knowledge-driven synthesis to dynamically generate follow-up questions and update the difficulty based on the estimations of target LLMs.
\textbf{(3) Evaluation Feedback}: Evaluating the target's deficiencies in multiple dimensions and providing actionable suggestions.
The overall framework is illustrated in Figure~\ref{fig: framework}, and detailed as pseudo code in Algorithm~\ref{pcode: judgeagent}.

\subsection{Benchmark Grading}
\label{subsec: 3_1_base_tiering}

To address the challenge of difficulty mismatch in prior dynamic evaluation, JudgeAgent first evaluates the target LLM on public static benchmarks and obtains a base score through a difficulty scoring mechanism. 
Similar to a written test before an interview, this allows JudgeAgent to form a basic capability estimation of the target LLM.

JudgeAgent first partitions the benchmark questions into batches.
Then it prompts the target LLM to answer the questions batch by batch. 
After assessing the responses, JudgeAgent estimates the target's capability based on its performance in each batch. 
Inspired by the common real-world practice of scoring student exams, we design a linear difficulty control mechanism. 
Within each batch, each correct answer improves the total score, and the average score determines the follow-up question's difficulty. 
Let $[d_1,d_2,d_3]$ respectively denote the difficulty \textit{Easy}, \textit{Medium}, and \textit{Hard}, 
for the score gain $c_{it}$ for a correct answer at different difficulty levels $d_i$, we set $[c_{1t},c_{2t},c_{3t}]=[1,1.5,2]$.
And for the threshold $t_{ij}$ between $d_i$ and $d_j$, we set $[t_{12},t_{23}]=[0.5,1]$.
Once the dynamic average score exceeds the threshold $t_{ij}$, the difficulty of the next generated question will be set to $d_j$.
The mechanism's detailed design, analysis, and proofs are available in Appendix~\ref{appendix: difficulty_control}.

\subsection{Interactive Extension}
\label{subsec: 3_2_interactive_extension}

This process is an "interview" where JudgeAgent conducts a knowledge-driven and difficulty-adaptive multi-turn evaluation.
Specifically, JudgeAgent dynamically adjusts the difficulty and generates questions with expanded knowledge based on the responses.
Furthermore, JudgeAgent explores the potential knowledge gaps of target LLMs through a multi-turn interaction process.
Each iteration in this process consists of three steps: Relevant Knowledge Retrieval, Difficulty-Adaptive Question Generation, and Capability Estimation.

\textbf{Relevant Knowledge Retrieval:}
To achieve knowledge-driven evaluation, JudgeAgent invokes external knowledge and constructs context graphs to facilitate the association and extension of knowledge during evaluation.
If only reference texts of seed questions are used for generation, the generated questions will be highly similar to the original ones.
Inspired by SoG\citep{sog}, JudgeAgent employs a context-graph–based sampling approach to get background texts. 
The context graph is constructed from the benchmark's knowledge base with entities and chunks extracted from texts as nodes. 
The entities in the same text will be linked on the graph.
The detailed construction is illustrated in Appendix~\ref{appendix: context_graph} and Algorithm~\ref{pcode: context_graph}.

During sampling, JudgeAgent first extracts entities from seed questions and finds the most similar entity $e$ on the graph.
In each hop, JudgeAgent retrieves the most relevant chunks of $e$ to the question, and randomly selects a chunk $c$. 
$e$ is the seed entity of the next hop, and $c$ is added to the path. 
After $N$ hops, we sample a path with breadth and depth, and the chunks on the path are concatenated to form the knowledge for generation. 
The context graph ensures both knowledge breadth and depth, and the greedy similarity strategy ensures relevance to seed questions. 
Entity extraction is performed by GPT-4.1, with prompts detailed in Appendix~\ref{appendix: prompts}. 

\textbf{Difficulty-Adaptive Question Generation:}
To mitigate data contamination and difficulty mismatch, JudgeAgent conducts multi-turn interactive interviews to explore potential deficiencies and incorporates adaptive difficulty adjustments into question generation.
A fixed-difficulty test can identify underperforming models, but can hardly comprehensively evaluate a model’s actual knowledge mastery. 
For instance, when an overly challenging test evaluates a model with limited knowledge, the evaluation may reveal that the model performs poorly, but fails to delineate the actual range of its knowledge. 
Therefore, JudgeAgent dynamically adjusts the follow-up question's difficulty based on capability estimation through the difficulty control mechanism in Section~\ref{subsec: 3_1_base_tiering}.

Based on the capability estimation, JudgeAgent adjusts the question difficulty as \textit{Easy}, \textit{Medium}, or \textit{Hard}, and explicitly specifies each difficulty's requirements in the prompt. 
For \textit{Easy}, the focus is on assessing information retrieval skills, primarily through cloze-style questions based on the original text, like \textit{"Which virus primarily causes HFMD?"}. 
For \textit{Medium}, the emphasis is on understanding key concepts, with questions involving basic inference, like \textit{"Which description is correct regarding meiosis II?"}
For \textit{Hard}, the questions are designed to encourage deep thinking and complex logical analysis, like \textit{"A patient ingested a toxic substance... The doctor employed... which type of treatment is likely lacking?"}.
Detailed prompt designs are provided in Appendix~\ref{appendix: prompts}, and the validation of question quality is described and analyzed in Appendix~\ref{appendix: question_quality}.

\textbf{Capability Estimation:}
JudgeAgent queries the target LLM with generated questions.
Based on LLM's responses, JudgeAgent computes average difficulty scores with the strategy in Section~\ref{subsec: 3_1_base_tiering}, and determines the difficulty for the next round's generation.
Through this adaptive multi-turn mechanism, JudgeAgent progressively aligns the assessment with the true capabilities of the target LLM, thereby achieving more precise evaluations.

\subsection{Evaluation Feedback}
\label{subsec: 3_3_evaluation_feedback}

Inspired by Generative Reward Model methods \citep{ankner2024critique, ye2024beyond, li2025test} and the evaluation report method in \citep{kim-etal-2025-llm-interviewer}, JudgeAgent generates a text evaluation report based on all Q\&A history for each batch.
Benefiting from context graphs, the evaluation report can identify the knowledge deficiencies of target LLMs and provide concise and valuable feedback around key knowledge concepts.
\section{Experiments}
\label{sec: experiment}

In this section, we validate the effectiveness of JudgeAgent using the method in Section~\ref{subsec: 3_3_evaluation_feedback}, which is detailed in Algorithm~\ref{pcode: validation}. 
The following research questions guide our experiments:
\textbf{RQ1}, does JudgeAgent's knowledge-driven evaluation genuinely discover the shortcomings of the target model? 
\textbf{RQ2}, is JudgeAgent's dynamic evaluation more effective and precise than the static benchmarking?
\textbf{RQ3}, to what extent does each mechanism in JudgeAgent influence the evaluations?

Current LLM knowledge evaluation methods lack quantifiable and standardized validation approaches. 
Therefore, to validate the effectiveness of JudgeAgent, we use a similar approach as \citep{kim-etal-2025-llm-interviewer}.
We use the evaluation report produced by JudgeAgent as suggestions to prompt the target LLM to answer the same questions again.
Then we compare the accuracy before and after the intervention to indirectly validate the effectiveness. 
To avoid cheating, we produce suggestions without directly providing the correct answer or background knowledge of seed questions to JudgeAgent.

\textbf{Experiment Setup.} We select GPT-4.1\footnote{We use gpt-4.1-2025-04-14 from OpenAI's official API for all experiments} to be the core LLM. In our experiments, the batch size in Benchmark Grading is 3, and the Interactive Extension is limited to a maximum of 3 rounds.

\subsection{Dataset and Target Model Selection}
\label{subsec: 5_1_dataset_model}


In our experiments, we select MedQA\citep{medqa},  MultiHop-RAG\citep{multihoprag}, and QuALITY\citep{quality} as the initial benchmarks.
We remove the background knowledge of questions from MedQA and MultiHop-RAG to evaluate the target's knowledge rather than its comprehension ability, and to verify whether JudgeAgent can discover the target's actual knowledge deficiencies.
We use QuALITY to validate the effectiveness in guiding comprehension and reasoning. The detailed information and the preprocessing procedures are in Appendix~\ref{appendix: dataset}.

For target LLMs, we select Qwen3\citep{qwen3}, GLM4-Flash\citep{chatglm}, GPT-4.1\citep{gpt}, and gemini-2.5-pro\citep{gemini}\footnote{We use qwen-plus-2025-04-28, gpt-4.1-2025-04-14, gemini-2.5-pro-preview-06-05, and the free version glm-4-flash-250414 as the target models.}. We utilize the official APIs to interact with these models.

\subsection{Evaluation Metrics}
\label{subsec: 5_2_metric}

The metrics used in our experiments are as follows:

\noindent(1) \textbf{Accuracy (ACC).} We use this metric to measure the performance of the target model on base datasets. 
To investigate the effectiveness of JudgeAgent, we compared the changes in the ACC before and after dynamic evaluation. ACC1 represents the target's ACC in answering base questions in \textit{Benchmark Grading}, measuring its performance before receiving evaluation feedback. ACC2 denotes the ACC in answering the same questions after receiving feedback from JudgeAgent.

\noindent(2) \textbf{Correction Rate (CR).} This metric quantifies the proportion of questions that the target model initially answered incorrectly but subsequently answered correctly after receiving evaluation suggestions, which measures the effectiveness of JudgeAgent. A higher correction rate indicates better performance of the suggestions.

\noindent(3) \textbf{Correct-to-Error Rate (CtE).} This metric serves as the inverse of the Correction Rate. A lower CtE indicates greater effectiveness.

\subsection{Main Results and Analysis}
\label{subsec: 5_3_main_result}

To address \textbf{RQ1}, we conduct experiments on three datasets to validate the effectiveness of JudgeAgent's evaluation, and the results are shown in Table~\ref{tb: exp_result_knowledge} and Table~\ref{tb: exp_result_quality}.

\begin{table*}[t]
\begin{minipage}{\linewidth}
    \begin{center}
    \footnotesize
    \begin{tabular}{l|cccc|cccc}
    \toprule
    \multirow{2}{*}{Target Model} & \multicolumn{4}{c|}{MedQA} & \multicolumn{4}{c}{MultiHopRAG} \\
     & ACC1 & ACC2 & CR$\uparrow$ & CtE$\downarrow$ & ACC1 & ACC2 & CR$\uparrow$ & CtE$\downarrow$ \\
    \midrule
    Qwen3 & 91.71 & 96.38 & 5.02 & 0.35 & 63.65 & 70.07 & 17.49 & 11.07 \\
    GLM4-Flash & 80.09 & 92.82 & 13.46 & 0.73 & 51.25 & 65.92 & 24.26 & 9.59 \\
    GPT-4.1 & 84.97 & 92.44 & 7.65 & 0.18 & 68.94 & 75.55 & 12.36 & 5.75 \\
    Gemini-2.5-pro & 91.04 & 94.60 & 4.03 & 0.47 & 62.25 & 71.83 & 15.41 & 5.83 \\
    \bottomrule
    \end{tabular}
    \end{center}
    \caption{The results on MedQA and MultiHopRAG, and all values are percentages.}
    \label{tb: exp_result_knowledge}
\end{minipage}

\vspace{1em}

\begin{minipage}{\linewidth}
    \begin{minipage}{\linewidth}
        \footnotesize
        \begin{center}
        \begin{tabular}{l|cccc|cccc}
        \toprule
        \multirow{2}{*}{Target Model} & \multicolumn{4}{c|}{QuALITY-overall} & \multicolumn{4}{c}{QuALITY-easy} \\
         & ACC1 & ACC2 & CR$\uparrow$ & CtE$\downarrow$ & ACC1 & ACC2 & CR$\uparrow$ & CtE$\downarrow$ \\
        \midrule
        Qwen3 & 87.83 & 88.84 & 1.88 & 0.87 & 94.63 & 95.61 & 0.98 & 0.00 \\
        GLM4-Flash & 73.48 & 76.38 & 4.78 & 1.88 & 83.26 & 84.65 & 4.19 & 2.79 \\ 
        GPT-4.1 & 89.13 & 92.75 & 3.91 & 0.29 & 93.66 & 96.10 & 2.44 & 0.00 \\
        Gemini-2.5-pro & 93.77 & 96.38 & 3.19 & 0.58 & 97.55 & 99.02 & 1.47 & 0.00 \\
        \bottomrule
        \end{tabular}
        \end{center}
    \end{minipage}

    \vspace{0.5em}
    
    \begin{minipage}{\linewidth}
        \footnotesize
        \begin{center}
        \begin{tabular}{l|cccc|cccc}
        \toprule
        \multirow{2}{*}{Target Model} & \multicolumn{4}{c|}{QuALITY-medium} & \multicolumn{4}{c}{QuALITY-hard} \\
         & ACC1 & ACC2 & CR$\uparrow$ & CtE$\downarrow$ & ACC1 & ACC2 & CR$\uparrow$ & CtE$\downarrow$ \\
        \midrule
        Qwen3 & 88.53 & 88.89 & 1.43 & 1.08 & 80.10 & 82.04 & 3.40 & 1.46 \\
        GLM4-Flash & 74.35 & 79.18 & 5.58 & 0.74 & 62.14 & 64.08 & 4.37 & 2.43 \\
        GPT-4.1 & 91.04 & 94.27 & 3.58 & 0.36 & 82.04 & 87.38 & 5.83 & 0.49 \\
        Gemini-2.5-pro & 93.57 & 95.36 & 2.50 & 0.71 & 90.29 & 95.15 & 5.83 & 0.97 \\
        \bottomrule
        \end{tabular}
        \end{center}
    \end{minipage}
    \caption{The results on QuALITY with different difficulty levels. QuALITY-X refers to the subdataset that consists of questions with specific difficulty, and -overall refers to all questions.}
    \label{tb: exp_result_quality}
\end{minipage}
\vspace{-1em}
\end{table*}

Based on the results from MedQA and MultiHopRAG in Table~\ref{tb: exp_result_knowledge}, JudgeAgent can effectively identify potential knowledge deficiencies in target LLMs and subsequently mitigate these deficiencies by providing targeted suggestions to the target LLM. 
Furthermore, as evidenced by the overall performance on QuALITY in Table~\ref{tb: exp_result_quality}, JudgeAgent also contributes to providing further guidance to address potential shortcomings in the logical reasoning abilities of the target LLM, thereby assisting in refining the target's thinking steps. 
Additionally, by comparing CR and CtE of different LLMs before and after receiving suggestions, it can be observed that the effectiveness of suggestions is less consistent for relatively weaker LLMs (e.g., the free model GLM4-Flash), as reflected in the higher CtE. 
In contrast, stronger LLMs are less susceptible to misleading suggestions.

Based on the performance across different difficulties of QuALITY in Table~\ref{tb: exp_result_quality}, we can analyze the effectiveness of JudgeAgent for different targets on various difficulty levels. 
For stronger LLMs (Qwen3, GPT-4.1, and Gemini-2.5-pro), JudgeAgent provides stronger guidance, particularly on questions of greater difficulty (\textit{Medium} and \textit{Hard}), since these LLMs may have already mastered the basic knowledge of \textit{Easy} questions. 
In contrast, for weaker LLMs, JudgeAgent leads to considerable improvement across all difficulty levels, with particularly notable gains on \textit{Easy} level, since JudgeAgent's feedback is more effective at filling gaps in basic concepts.
These results further indicate that JudgeAgent more accurately assesses the capability and provides more difficulty-adaptive guidance to the target LLMs. 
Furthermore, for all target LLMs, JudgeAgent's suggestions demonstrate clear guidance on \textit{Hard} questions, indicating that JudgeAgent effectively identifies underlying deficiencies in target LLMs through dynamic interactive evaluation.

In summary, JudgeAgent can effectively and precisely identify potential knowledge deficiencies in target LLMs, offering more refined evaluation results.
Furthermore, we conducted cross-validation experiments to verify that the JudgeAgent's evaluations are effective not only for seed questions, which are detailed in Appendix~\ref{appendix: cross_validation}.
Additionally, experiments on the efficiency and stability of JudgeAgent are detailed in Appendix~\ref{appendix: efficiency} and~\ref{appendix: stability}.

\subsection{Analysis of JudgeAgent's Dynamic Scores}
\label{subsec: 5_4_score}

\begin{table*}[t]
\begin{center}
\footnotesize
\begin{tabular}{c|l|c|cccc|ccc}
\toprule
\multirow{2}{*}{\textbf{Benchmark}} & \multirow{2}{*}{\textbf{Target Model}} & \multirow{2}{*}{\textbf{ACC(\%)}} & \multicolumn{4}{c|}{\textbf{Difficulty Score}} & \multicolumn{3}{c}{\textbf{Difficulty Distribution(\%)}} \\
 & & & base & @1 & @2 & @3 & easy & medium & hard \\
\midrule
\multirow{4}{*}{MedQA} & GLM4-Flash & 80.09 & 1.201 & 1.230 & 1.217 & 1.192 & 20.86 & 12.64 & 66.50 \\
 & GPT-4.1 & 84.97 & 1.275 & 1.388 & 1.395 & 1.401 & 14.61 & 9.02 & 76.38 \\
 & Qwen3 & 91.71 & 1.376 & 1.472 & 1.496 & 1.487 & 8.01 & 8.58 & 83.42 \\
 & Gemini-2.5-pro & 91.04 & 1.366 & 1.484 & 1.494 & 1.505 & 9.41 & 7.76 & 82.84 \\
\midrule
\multirow{4}{*}{MultiHopRAG} & GLM4-Flash & 51.25 & 0.769 & 0.587 & 0.535 & 0.486 & 34.06 & 48.13 & 17.81 \\
 & GPT-4.1 & 68.94 & 1.034 & 1.006 & 1.000 & 0.996 & 11.55 & 40.16 & 48.29 \\
 & Qwen3 & 63.65 & 0.955 & 0.801 & 0.767 & 0.746 & 19.62 & 48.91 & 31.47 \\
 & Gemini-2.5-pro & 62.25 & 0.934 & 0.967 & 1.005 & 1.019 & 13.63 & 38.80 & 47.57 \\
\midrule
\multirow{4}{*}{QuALITY} & GLM4-Flash & 74.35 & 1.065 & 1.143 & 1.232 & 1.304 & 12.09 & 39.30 & 48.60 \\
 & GPT-4.1 & 91.04 & 1.320 & 1.428 & 1.522 & 1.603 & 1.86 & 23.49 & 74.65 \\
 & Qwen3 & 88.53 & 1.296 & 1.388 & 1.466 & 1.520 & 3.26 & 25.58 & 71.76 \\
 & Gemini-2.5-pro & 93.57 & 1.397 & 1.502 & 1.595 & 1.686 & 0.00 & 17.91 & 82.09 \\
\bottomrule
\end{tabular}
\end{center}
\caption{The dynamic difficulty scores and difficulty distribution of questions generated by JudgeAgent. \textit{base} means the score for base static benchmark questions, and \textit{@K} means the scores for extended questions at the K-th iteration.}
\label{tb: exp_result_difficulty_score}
\vspace{-1em}
\end{table*}

To address \textbf{RQ2}, we analyze the dynamic difficulty scores and the difficulty distribution of generated questions during JudgeAgent’s Interactive Expansion stage.
The results are shown in Table~\ref{tb: exp_result_difficulty_score}.

As the results on MedQA show, there is no significant difference in the accuracy and scores between GPT-4.1 and GLM4-Flash on the static benchmark (\textit{base}).
However, through multi-round dynamic evaluation, it can be observed that the difficulty of questions queried to GPT-4.1 continuously increases, while that for GLM4-Flash decreases, indicating that GPT-4.1's actual knowledge mastery exceeds the preset difficulty of the static benchmark, further suggesting that GPT-4.1's knowledge mastery is far higher than that of GLM4-Flash.

Similarly, on MultiHopRAG, Qwen3 and Gemini-2.5-pro show comparable performance on the static benchmark. 
While in dynamic evaluations, Qwen3's performance is significantly lower than that of GPT-4.1 and Gemini-2.5-pro, with the gap progressively widening over rounds.
This gap is also reflected in the fact that the generated questions for Qwen3 are primarily in \textit{Medium} level. 
Meanwhile, although GPT-4.1 performs noticeably better than Gemini-2.5-pro on the static benchmark, their performance was comparable in dynamic evaluations, as can be seen from their similar difficulty distributions in the questions generated.
Moreover, the multi-turn interview mechanism makes the differentiation in evaluation more pronounced.
In summary, compared to static benchmarking, JudgeAgent's dynamic evaluation can more precisely distinguish the actual knowledge mastery of different LLMs.
Furthermore, we validate JudgeAgent's resistance to data contamination through dynamic evaluation, with experiments and analysis detailed in Appendix~\ref{appendix: exp_data_contamination}.

\subsection{Ablation Study}
\label{subsec: 5_5_ablation}

\begin{figure}[ht]
    \centering
    \includegraphics[width=\linewidth]{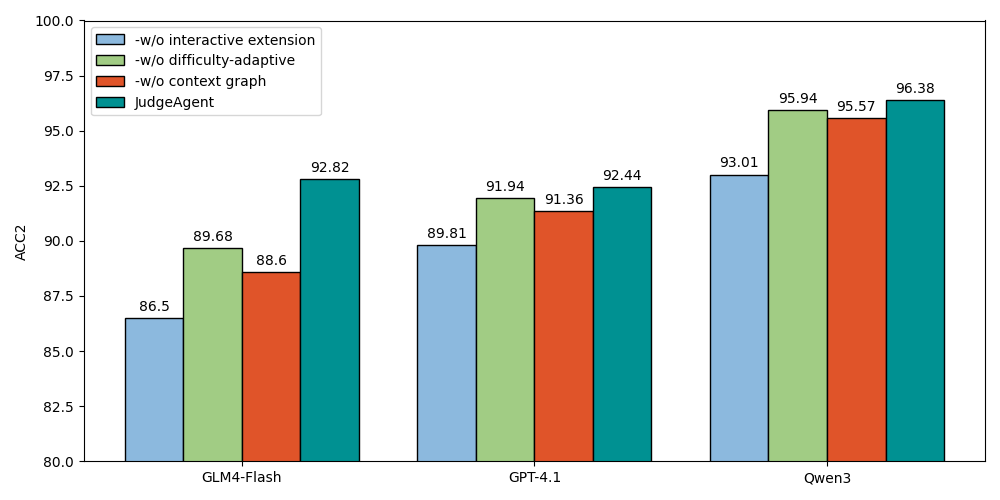}
    \caption{The results of the ablation study with MedQA as the base dataset, and all values are percentages.}
    \label{fig: ablation_study}
    \vspace{-1em}
\end{figure}

To address \textbf{RQ3}, we conduct ablation studies on MedQA with GLM4-Flash as the target LLM. 
When maintaining the responses to seed questions unchanged, we removed different modules of JudgeAgnet. 
By comparing the results under different ablation settings, we investigate how much different modules influence the evaluations
The results are shown in Figure~\ref{fig: ablation_study}.

\begin{figure*}[t]
    \centering
    \includegraphics[width=\linewidth]{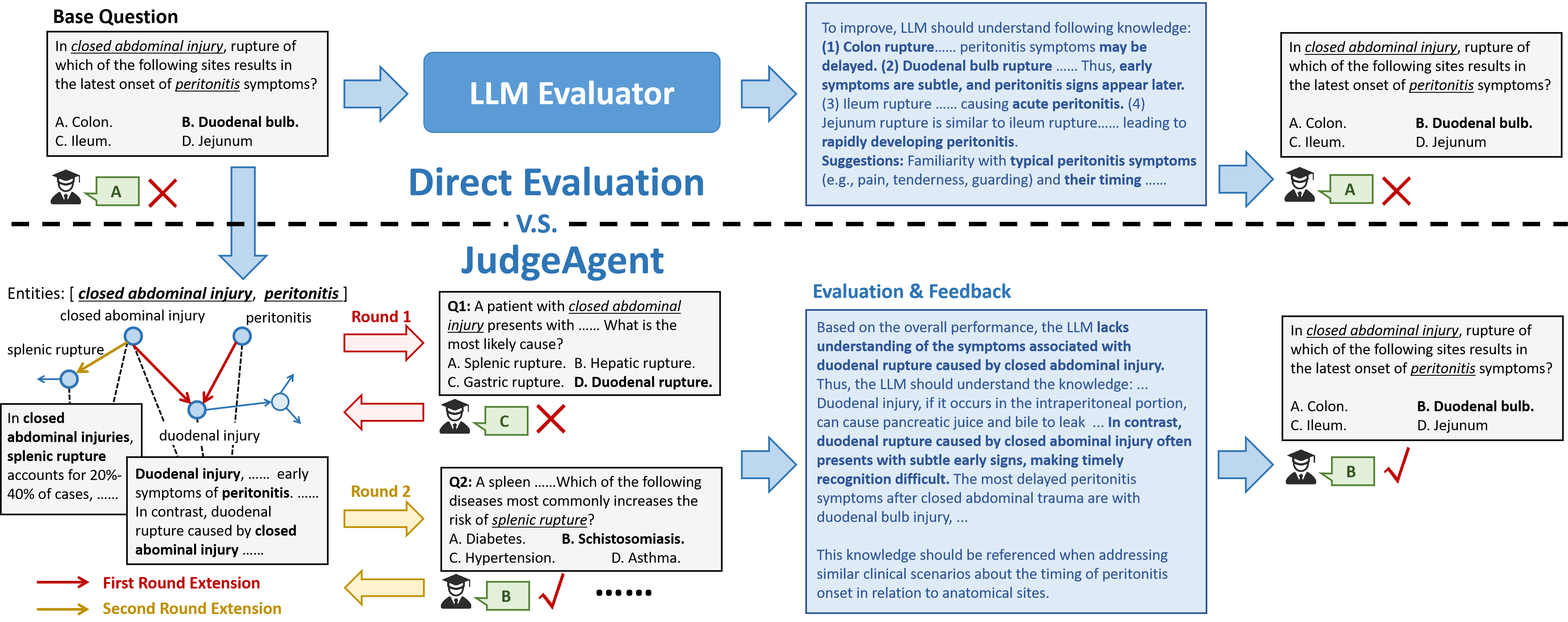}
    \caption{The brief overview of the comparative case in the Case Study.}
    \label{fig: case_study_overview}
    \vspace{-1em}
\end{figure*}


JudgeAgent (\textit{w/o} context graph) removes the context graph and only uses chunks sampled randomly from the knowledge base. 
The results indicate that removing the context graph reduces the effectiveness of the JudgeAgent. 
The lack of context graph severs the knowledge link between extended questions and base ones, leading to feedback without accurate information, which may disrupt the thinking of target LLMs.
The higher CtE compared to the setting -\textit{w/o} difficulty-adaptive also provides supporting evidence for this potential interference.

JudgeAgent (\textit{w/o} difficulty-adaptive) removes the difficulty-adaptive mechanism and generates questions with fixed difficulty rules in the prompt. 
The results show that the removal of this module diminishes the effectiveness of JudgeAgent’s evaluations, demonstrating the importance of the difficulty-adaptive mechanism for providing effective evaluations.

JudgeAgent (\textit{w/o} interactive extension) removes the Interactive Extension and only evaluates models with base benchmarks. 
The results show that removing this stage significantly weakens the effectiveness of JudgeAgent compared to other ablation settings,
indicating that the dynamic expansion, which expands both breadth and depth of knowledge, is crucial to the evaluation's effectiveness. 


\subsection{Parameter Analysis}
\label{subsec: 5_6_parameter}


\begin{figure}[ht]
    \centering
    \includegraphics[width=\linewidth]{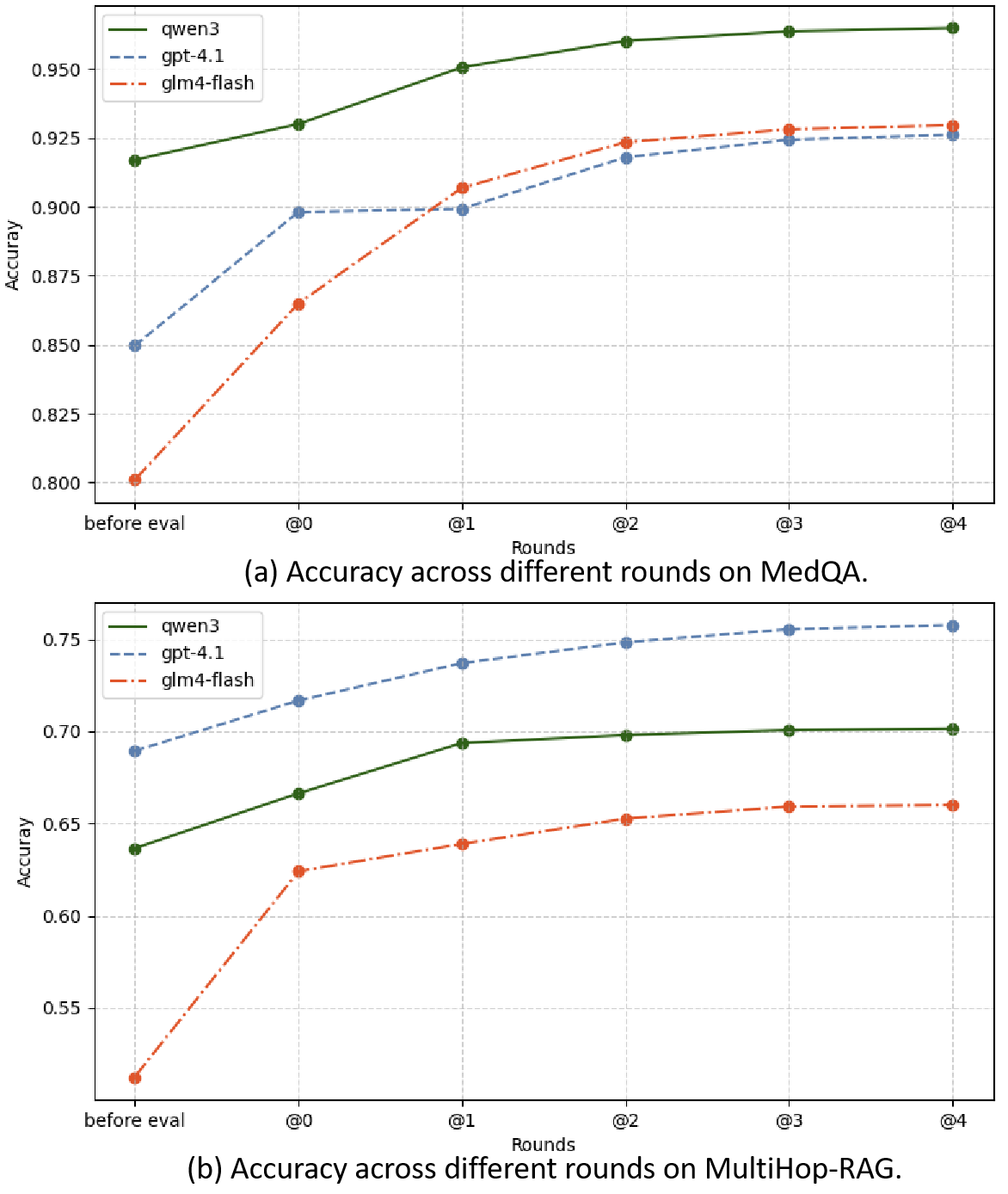}
    \caption{The results of different expansion rounds on MedQA and MultiHop-RAG. @K indicates the ACC improvement after the K-th interaction.}
    \label{fig: parameter_analysis}
    \vspace{-1em}
\end{figure}

The number of extension rounds is crucial for balancing the efficiency and effectiveness of JudgeAgent. 
We test the JudgeAgent's effectiveness under different rounds in Interactive Extension, as shown in Figure~\ref{fig: parameter_analysis}.
The results demonstrate that as the rounds increase, the JudgeAgent's effectiveness, which is represented by the accuracy improvement of target LLMs, also gradually improves. 
However, the trend gradually slows down, showing a relatively clear marginal effect. 

As the rounds increase, JudgeAgent can expand more questions related to the knowledge around base questions. 
After reaching certain rounds, the Q\&A history is sufficient to identify the target's knowledge deficiencies around base questions.
Additional questions serve only as corroborative rather than critical evidence.
There is a clear marginal effect in Figure~\ref{fig: parameter_analysis}. 
For example, the results of GPT-4.1 on MultiHop-RAG, the improvement per round drops from 2.04\% to 1.13\%, then to 0.71\%, and eventually to 0.23\%, and other curves exhibit similar patterns.
Therefore, JudgeAgent expands a maximum of 3 rounds in our experiments to avoid wasting resources and time.

Additionally, parameter analysis experiments regarding batch size are presented in Appendix~\ref{appendix: batch_size_analysis}.

\subsection{Case Study}
\label{subsec: 5_7_case}


To further understand JudgeAgent's mechanism, we analyze GLM4-Flash's responses to MedQA questions by receiving suggestions from direct evaluation and JudgeAgent. 
The seed questions where the target answers correctly are omitted in this case.
A brief overview of the case is shown in Figure~\ref{fig: case_study_overview}, and the detailed content is provided in Appendix~\ref{appendix: case_study}.

In this case, the target LLM answers the base question incorrectly. 
When directly evaluated, the evaluator LLM generically enumerated the potential consequences of closed abdominal injury. 
Consequently, even after receiving the feedback, the target LLM still outputs an incorrect answer.

In contrast, JudgeAgent first extracted two key entities, \textit{closed abdominal injury} and \textit{peritonitis}.
Based on the sampled paths on the context graph, JudgeAgent generated extended questions, which help discover the lack of understanding of \textit{duodenal injury} and \textit{closed abdominal injury}.
Ultimately, the target LLM successfully answered the original question correctly with this feedback.


\section{Conclusion}
\label{sec: conclusion}

In this paper, we propose JudgeAgent, a knowledge-driven and dynamic evaluation framework for LLMs. 
To address the challenge of limited knowledge coverage, JudgeAgent leverages LLM agents with context graphs to traverse knowledge structures and employs knowledge-driven synthesis for question generation.
Furthermore, to mitigate data contamination and difficulty mismatch, JudgeAgent dynamically adjusts the capability estimation and generates difficulty-adaptive questions based on the target's responses in multi-turn interviews.
Extensive experiments validate the effectiveness of JudgeAgent and highlight the significant potential of this knowledge-driven and dynamic evaluation paradigm in enhancing the evaluation results.
In our future work, we will further refine this dynamic evaluation paradigm and develop more effective and reliable evaluation tools.

\section*{Limitations}
\label{sec:limitations}

In this paper, we validate JudgeAgent's effectiveness and its great potential for enhancing the dynamic evaluation paradigm of LLMs through extensive experiments. 
However, due to space constraints, we focus on proposing a knowledge-driven and dynamic evaluation framework, and do not delve deeply into methods for optimizing the evaluated LLMs based on JudgeAgent. 
In Section~\ref{sec: experiment} Experiments, we only employ simple prompt injection optimization as a validation approach to verify the evaluation's effectiveness. 
In future work, we plan to further explore the potential of this evaluation paradigm, conducting in-depth research on how to perform targeted training and optimization of the evaluated LLMs based on JudgeAgent’s evaluation results, thereby establishing an efficient and comprehensive iteration cycle for LLMs.
\section*{Ethical Considerations}
\label{sec: ethics_statement}
In this study, all datasets used, including MedQA, MultiHop-RAG, and QuALITY, were sourced in compliance with relevant usage guidelines, ensuring no violation of privacy. No personally identifiable information was used, and no experiments were conducted that could raise privacy or security concerns. We are committed to maintaining transparency and integrity throughout the research process. 

\bibliography{custom}

\appendix
\section{The Use of Large Language Models}
\label{appendix: use_llms}
In this paper, we only used LLMs, including DeepSeek-R1 and ChatGPT, for polishing the writing.
Specifically, we used LLMs to assist in refining the language, improving readability, and grammar checking. The authors take full responsibility for the content of the paper.
\section{Analysis and Proofs of Adaptive Difficulty-Control Mechanisms}
\label{appendix: difficulty_control}

In Section~\ref{subsec: 3_1_base_tiering}, we have formalized the difficulty control mechanisms as follows:

Let $[d_1,d_2,d_3]$ respectively denote the difficulty \textit{Easy}, \textit{Medium}, and \textit{Hard}, $n_{it},n_{if}$ respectively represent the number of correct and incorrect answers for $d_i$-level questions, $c_{it}$ be the score gain for a correct answer at $d_i$, and $t_{ij}$ be the score threshold between $d_i$ and $d_j$, we can get the average score $avg_s$ and the difficulty of the next round question generation $d_{next}$:
\begin{equation*}
    \begin{aligned}
        avg_{s}&=\frac{\sum_{i=1}^3c_{it}n_{it}}{\sum_{i=1}^3(n_{it}+n_{if})}\\
        d_{next}&=\left\{
        \begin{aligned}
            &d_1,\quad avg_{s}\leq t_{12}\\
            &d_2,\quad avg_{s}\leq t_{23}\\
            &d_3,\quad avg_{s}>t_{23}\\
        \end{aligned}
        \right.
    \end{aligned}
\end{equation*}
where $[c_{1t},c_{2t},c_{3t}]=[1,1.5,2]$ and $[t_{12},t_{23}]=[0.5,1]$ in our experiments.
In this section, we will illustrate the criteria and rules of the score gain $c_{it}$ and the threshold $t_{ij}$, along with the analysis and proofs.
For ease of representation, let $N=\sum_{i=1}^3(n_{it}+n_{if})$ in the subsequent analysis.

For the selection of score gains and thresholds, we followed simple cognitive principles to establish several rules that a reasonable difficulty control mechanism should satisfy: 

\textbf{R1. Balance:} Correctly answering a more difficult question should gain a higher score. If the number of correctly answered \textit{Easy} and \textit{Hard} questions is equal, the level should be considered equivalent to the same total number of \textit{Medium} questions. Additionally, the conventional range of the average score should correspond equally to the three difficulty levels in ascending order.

\textbf{R2. Generalizability:} The same formula, score gains, and thresholds should apply even to questions without difficulty labels, where difficulty is estimated solely based on the accuracy.

\textbf{R3. Improvability:} Correctly answering questions should increase the average score, and there should be a possibility of exceeding the threshold to advance to the next difficulty level.

\textbf{R4. Stability:} If the capability estimation remains at a certain difficulty for a long period, the likelihood of advancing to the next level should gradually decrease.

Following the above rules, we can analyze the mathematical conditions that $c_{it}$ and $t_{ij}$ must satisfy.
From \textbf{R1}, we can get that $c_{1t}+c_{3t}=2c_{2t}\land c_{1t}<c_{2t}<c_{3t}$. 
Since we have set the score for correctly answering a question without a difficulty level as $c_{2t}$, the range of $avg_s$ can be determined as $[0\,,\,c_{2t}]$, representing the spectrum from answering all questions incorrectly to answering all correctly.
Based on the condition about threshold in \textbf{R1}, we can get that $t_{12}=c_{2t}/3\ ,\ t_{23}=2c_{2t}/3$ in the case that difficulty levels are absent. 
Due to \textbf{R2}, these threshold conditions can be generalized to the case where questions are provided with difficulty levels.

For \textbf{R3}, let $avg_s$ denote the average score after answering $N$ questions, and $avg_s'$ denote the average score after answering $N+1$ questions.
Then \textbf{R3} is equivalent to satisfying the following conditions:
\begin{equation}
    \begin{aligned}
        &\textbf{C1.}\quad avg_s'=\frac{avg_sN+c_{it}}{N+1}>avg_s,\\
        &\quad\text{holds true always when}\ \ avg_s\leq t_{(i,i+1)}\\\\
        &\textbf{C2.}\quad\exists\ \ avg_{s0},\, N_0,\\
        &\quad\text{s.t.}\ avg_s'=\frac{avg_{s0}N_0+c_{it}}{N_0+1}>t_{(i,i+1)}
    \end{aligned}
    \label{eq: condition_r3}
\end{equation}

Simplifying Eq~\ref{eq: condition_r3}.C1, we find that $c_{it}>avg_s$ holds true always when $avg_s\leq t_{(i,i+1)}$. Therefore, we can conclude that $c_{it}>t_{(i,i+1)}$, meaning $c_{2t}>c_{1t}>t_{12}$ and $t_{23}<c_{2t}<c_{3t}$. 
Simplifying Eq~\ref{eq: condition_r3}.C2, we can obtain a relationship between $avg_{s0}$ and $N_0$: 
\begin{equation}
    avg_{s0}>t_{(i,i+1)}-\frac{c_{it}-t_{(i,i+1)}}{N_0}
\end{equation}
Combining the condition $c_{it}>t_{(i,i+1)}$ from Eq~\ref{eq: condition_r3}.C1, this relationship indicates that, under the premise of \textbf{R3}, advancing to the next difficulty level by answering questions correctly requires a higher $avg_s$ as $N$ increases. 
This description is essentially \textbf{R4}: the longer one remains at a specific difficulty level, the harder it becomes to advance to the next level. 
Therefore, \textbf{R3}$\Rightarrow$\textbf{R4} is true.

Finally, \textbf{R4} can be stated as follows: given the average score $a\leq t_{(i,i+1)}$ after $N$ questions, the number $n$ of consecutive questions that must be answered correctly to surpass the level threshold should increase with $N$. We can obtain the following inequality:
\begin{equation}
    \begin{aligned}
        &\frac{aN+c_{it}n}{N+n}>t_{(i,i+1)}\\
        &\quad\Rightarrow(c_{it}-t_{(i,i+1)})n>(t_{i,i+1}-a)N
    \end{aligned}
\end{equation}
If \textbf{R4} holds, which means $n$ should increase as $N$ increases, combined with the inherent condition $a\leq t_{(i,i+1)}$, it follows that $c_{it}>t_{(i,i+1)}$, from which R3 can be deduced. 
Therefore, \textbf{R3}$\Rightarrow$\textbf{R4} and \textbf{R4}$\Rightarrow$\textbf{R3} hold simultaneously, meaning \textbf{R4} and \textbf{R3} are equivalent, demonstrating that we have now analyzed all the mathematical conditions that $c_{it}$ and $t_{(i,i+1)}$ must satisfy:
\begin{equation}
    \begin{aligned}
    &c_{1t}<c_{2t}<c_{3t}\\
    &c_{1t}+c_{3t}=2c_{2t}\\
    &t_{12}=c_{2t}/3\ ,\ t_{23}=2c_{2t}/3\\
    &c_{it}>t_{(i,i+1)}\\
    &\qquad\Rightarrow c_{2t}>c_{1t}>t_{12}\ ,\ t_{23}<c_{2t}<c_{3t}
    \end{aligned}
\end{equation}

Therefor, we first set $c_{2t}=1.5$, and so $t_{12}=0.5\ ,\ t_{23}=1$. Then, since $1.5=c_{2t}>c_{1t}>t_{12}=0.5$ and $1=t_{23}<c_{2t}=1.5<c_{3t}$, we set $c_{1t}=1$ for convenience, and so $c_{3t}=2c_{2t}-c_{1t}=2$.

\section{Construction of Context Graph}
\label{appendix: context_graph}

For the dataset selected during the Benchmark Grading stage, the process of constructing a context graph from its knowledge base (e.g., the set of all reference texts) can be divided into three components: text chunking, node construction, and node linking.

\textbf{Text Chunking:}
To preserve the integrity of knowledge, the granularity of chunking is not refined to the sentence level. Instead, several sentences or an entire paragraph are grouped to form a single unit, treated as the minimal hierarchical “chunk”. On this basis, chunks are further aggregated into higher-level “documents” according to whether they belong to the same article or serve as reference texts for the same question.

\textbf{Node Construction:}
An LLM is employed to extract entities from each chunk, and the detailed prompt is provided in Appendix~\ref{appendix: prompts}. Each extracted entity serves as the subject for constructing a node. All chunks containing the same entity are assigned to the corresponding entity node. Formally, given an entity $e$, along with the chunks containing the entity $\{c_1, c_2,...\}$ and the documents containing the entity $\{d_1,d_2,...\}$, the node in the context graph can be defined as follows:
\begin{equation}
    \begin{aligned}
        &N=(e\,,\,\mathcal{C}\,,\,\mathcal{D})\ ,\\ 
        &\quad\text{where}\ \mathcal{C}=\{c_1,c_2...\}\ ,\ \mathcal{D}=\{d_1,d_2,...\}
    \end{aligned}
\end{equation}
In practice, we store only the IDs of chunks and documents within each node to conserve space.

\textbf{Node Linking:}
During construction, if two entities appear together in the same chunk or the same document-level texts, their corresponding nodes in the context graph are treated as neighbors. Formally, given two nodes $N_1=(e_1\,,\,\mathcal{C}_1\,,\,\mathcal{D}_1)$ and $N_2=(e_2\,,\,\mathcal{C}_2\,,\,\mathcal{D}_2)$, if $\exists\ c_{1i}\in\mathcal{C}_1, c_{2j}\in\mathcal{C}_2\,,\,\text{s.t.}\ c_{1i}\equiv c_{2j}$ or $\exists\ d_{1i}\in\mathcal{D}_1, d_{2j}\in\mathcal{D_2}\,,\,\text{s.t.}\ d_{1i}\equiv d_{2j}$, then $N_1$ and $N_2$ will be treated as neighbors in the context graph.
\section{Statistics and Preprocessing of Datasets}
\label{appendix: dataset}

\subsection{Details of Datasets}
\label{appendix: dataset_details}

\begin{table*}[ht]
\begin{center}
\begin{tabular}{llllll}
\toprule
\textbf{Datasets} & \textbf{Question Type} & \textbf{Categories} & \textbf{Splits} & \textbf{Used Splits} & \textbf{Split Size}\\
\midrule
\multirow{3}{*}{MedQA} & \multirow{3}{*}{multiple-choice} & medical & train & \multirow{3}{*}{test} & \multirow{3}{*}{1273} \\
 & & clinical & validation & & \\
 & & & test & & \\
\midrule
\multirow{6}{*}{MultiHopRAG} & \multirow{6}{*}{phrase QA} & technology & \multirow{6}{*}{test} & \multirow{6}{*}{test} & \multirow{6}{*}{2556} \\
 & & entertainment & & & \\
 & & sports & & & \\
 & & science & & & \\
 & & business & & & \\
 & & health & & & \\
\midrule
\multirow{3}{*}{QuALITY} & \multirow{3}{*}{multiple-choice} & fiction stories & train & \multirow{3}{*}{validation} & \multirow{3}{*}{2086} \\
 & & magazine articles & validation & & \\
 & & long articles & test & & \\
\bottomrule
\end{tabular}
\end{center}
\caption{Details of datasets in our experiments. The language of all the datasets is English.}
\label{tb: dataset}
\vspace{-5pt}
\end{table*}

The following benchmarks are used in our experiments, whose details are shown in Table~\ref{tb: dataset}.

\textbf{MedQA}\citep{medqa} contains multiple-choice questions in the style of the Medical Licensing Examination.
Questions in this dataset are collected from medical board exams in the US, Mainland China, and Taiwan, where human doctors are evaluated on their professional knowledge and ability to make clinical decisions.
The background knowledge texts of MedQA are provided in the form of additional complete articles, and the questions only provide meta information.

\textbf{MultiHop-RAG}\citep{multihoprag} consists of phrase Q\&A queries, their ground truth answers, and the associated supported evidence constructed from news articles published between September and December 2023.
The background knowledge texts of MultiHop-RAG are provided as supporting evidence along with the questions.

\textbf{QuALITY}\citep{quality} is a multiple-choice question dataset for long document comprehension, whose questions are written and validated by human contributors based on the long passages.
The sources of QuALITY include: (1) Project Gutenberg fiction stories, which are mostly science fiction; (2) Slate magazine articles from the Open American National Corpus; (3) other nonfiction articles taken from The Long+Shor, Freesouls, and the book Open Access.
QuALITY is organized by articles, with each data item consisting of a long article and several related questions, and the article serves as the background knowledge for each question.

\subsection{Preprocessing of Datasets}
\label{appendix: dataset_preprocess}

To verify JudgeAgent's ability to evaluate knowledge deficiencies in target models, we remove the background knowledge of the questions from MedQA and MultiHopRAG during evaluation. 
Otherwise, it would only test the target's reading comprehension ability rather than knowledge deficiencies.
In contrast, we provided the background text when using QuALITY for evaluation, as the questions in QuALITY are highly dependent on the text content, and many of the texts are fictional narratives.
Therefore, we use QuALITY to validate the JudgeAgent’s effectiveness in guiding comprehension and reasoning rather than discovering knowledge deficiencies.

Additionally, we have reprocessed the difficulty levels of the questions from QuALITY. 
Based on the accuracy of human annotators in answering questions, QuALITY originally classified questions into two levels, \textit{Easy} and \textit{hard}, using a 50\% accuracy threshold. 
To align with the difficulty levels defined by JudgeAgent's difficulty control module (\textit{Easy}, \textit{Medium}, and \textit{Hard}), we re-labeled the questions using the same accuracy criteria. 
Questions of QuALITY with an accuracy below 1/3 are re-labeled as \textit{Easy}, those with an accuracy between 1/3 and 2/3 as \textit{Medium}, and the rest as \textit{Hard}.
\section{Additional Experiment Analysis}
\label{appendix: additional_experiment}

\subsection{Efficiency of JudgeAgent}
\label{appendix: efficiency}

\begin{table}[h]
    \centering
    \footnotesize
    \setlength{\tabcolsep}{3pt}
    \resizebox{\linewidth}{!}{
        \begin{tabular}{l|cc|cc}
            \toprule
            \multirow{2}{*}{\textbf{Target}} & \multicolumn{2}{c|}{\textbf{MedQA}} & \multicolumn{2}{c}{\textbf{MultiHopRAG}} \\
             & avg.token & avg.time & avg.token & avg.time \\
            \midrule
            GLM4-Flash & 1558 & 7.24 & 2424 & 23.38 \\
            Qwen3 & 1909 & 9.95 & 3453 & 30.96 \\
            \bottomrule
        \end{tabular}
    }
    \caption{Efficiency of JudgeAgent. \textit{avg.tokens} and \textit{avg.time} respectively denote the average tokens and time (seconds) cost of JudgeAgent per seed question.}
    \label{tab: time_token_cost}
    \vspace{-5pt}
\end{table}

We calculated the time and token costs incurred by JudgeAgent during multi-turn interactions with GLM4-Flash and Qwen3 as target models, calculated as averages based on the number of questions in the benchmark selected during the Benchmark Grading stage. 
The results are shown in Table~\ref{tab: time_token_cost}.

\subsection{Stability of JudgeAgent}
\label{appendix: stability}

\begin{table*}[ht]
\begin{minipage}{\linewidth}
    \centering
    \begin{tabular}{l|cccc}
        \toprule
        \multirow{2}{*}{\textbf{Target}} & \multicolumn{4}{c}{\textbf{MedQA}} \\
         & ACC1 & ACC2 & CR$\uparrow$ & CtE$\downarrow$ \\
        \midrule
        Qwen3 & 91.76$\pm$0.09 & 96.18$\pm$0.17 & 4.83$\pm$0.17 & 0.42$\pm$0.08 \\
        GLM4-Flash & 80.36$\pm$0.34 & 92.65$\pm$0.29 & 13.00$\pm$0.38 & 0.72$\pm$0.18 \\
        \bottomrule
    \end{tabular}
\end{minipage}

\vspace{1em}

\begin{minipage}{\linewidth}
    \centering
    \begin{tabular}{l|cccc}
        \toprule
        \multirow{2}{*}{\textbf{Target}} & \multicolumn{4}{c}{\textbf{MultiHopRAG}} \\
         & ACC1 & ACC2 & CR$\uparrow$ & CtE$\downarrow$ \\
        \midrule
        Qwen3 & 63.97$\pm$0.53 & 70.41$\pm$0.59 & 15.78$\pm$1.29 & 9.34$\pm$1.35 \\
        GLM4-Flash & 51.54$\pm$0.38 & 66.56$\pm$0.51 & 23.44$\pm$0.88 & 8.22$\pm$1.03 \\
        \bottomrule
    \end{tabular}
\end{minipage}
\caption{The experiment results with standard deviation calculated based on 5 different random seeds.}
\label{tab: result_stability}
\end{table*}

To investigate the stability of JudgeAgent’s results, this section presents additional experiments conducted under 5 different random seeds, with the standard deviation calculated. 
Experiments are conducted using Qwen3 and GLM4-Flash as target models on the MedQA and MultiHop-RAG benchmarks. 
The results are shown in Table~\ref{tab: result_stability}.

The experiment results demonstrate that the results across multiple random seeds remain stable and are consistent with the ranking of results presented in the main results in Table~\ref{tb: exp_result_knowledge}. 
This stability benefits from JudgeAgent's context-graph control and knowledge-driven synthesis pathway for evaluation questions. 
These mechanisms endow the generated questions and the evaluation feedback with controllability and stability.

\subsection{The Resilience Against Data Contamination}
\label{appendix: exp_data_contamination}
\textbf{Can JudgeAgent mitigate the challenges of data contamination in static benchmarking paradigms?}
In this section, we simulate a scenario where the static benchmarking evaluation paradigm suffers from data contamination by deliberately exposing the evaluation questions to LLMs during their training process. 
We selected Llama3-8B-Instruct, Mistral-7B-Instruct-v0.3, and Qwen2.5-7B-Instruct as base models, and constructed supervised fine-tuning (SFT) data from the MedQA and MultiHop-RAG benchmarks, which are intended for evaluation in the main experiments. 
These training data were used to fine-tune the selected base models. 
By comparing the performance differences between the original base models and the fine-tuned models on both the static benchmark questions and the extended questions generated by JudgeAgent, we analyze and verify the resilience of JudgeAgent and its derivative JudgeAgent against data contamination. 

\begin{table*}[t]
\begin{center}
\begin{tabular}{c|cccc|cccc}
\toprule
\multirow{2}{*}{Models} & \multicolumn{4}{c|}{MedQA} & \multicolumn{4}{c}{MultiHop-RAG} \\
 & base & @1 & @2 & @3 & base & @1 & @2 & @3 \\
\midrule
Llama3-8B-Instruct & 62.17 & 69.94 & 67.72 & 69.77 & 48.77 & 28.30 & 28.38 & 29.15 \\
\hfill Llama3-sft & 82.53 & 65.59 & 65.80 & 65.42 & 88.45 & 29.59 & 29.19 & 29.79 \\
\hfill$\Delta$ & \textbf{20.34} & -4.35 & -1.92 & -4.35 & \textbf{39.69} & 1.29 & 0.81 & 0.65 \\
\midrule
Mistral-7B-Instruct-v0.3 & 57.91 & 59.40 & 58.76 & 56.97 & 59.51 & 37.75 & 38.60 & 37.22 \\
\hfill Mistral-sft & 62.73 & 53.18 & 51.98 & 51.98 & 92.17 & 32.74 & 32.58 & 32.62 \\
\hfill$\Delta$ & \textbf{4.82} & -6.23 & -6.78 & -4.99 & \textbf{32.66} & -5.01 & -6.02 & -4.60 \\
\midrule
Qwen2.5-7B-Instruct & 85.93 & 78.04 & 75.44 & 77.31 & 46.31 & 20.59 & 21.64 & 21.48 \\
\hfill Qwen2.5-sft\quad & 94.20 & 78.00 & 75.78 & 77.27 & 88.90 & 26.60 & 25.56 & 24.75 \\
\hfill$\Delta$ & \textbf{8.27} & -0.04 & 0.34 & -0.04 & \textbf{42.59} & 6.02 & 3.92 & 3.27 \\
\bottomrule
\end{tabular}
\end{center}
\caption{The results for validating the resilience against data contamination. \textit{base} means the performance on base static benchmark questions, and \textit{@K} means extended questions at the K-th iteration.}
\label{tb: exp_result_data_contamination}
\end{table*}

Specifically, the procedure can be summarized as follows: for a base LLM $\mathcal{M}$ and a static evaluation benchmark $\mathcal{D}$, a fine-tuned LLM $\mathcal{M}$-sft is obtained by supervised fine-tuning with the training data constructed from $\mathcal{D}$. 
The performance of $\mathcal{M}$ and $\mathcal{M}$-sft, which is measured by the accuracy (ACC) in answering the questions, is then compared on both $\mathcal{D}$ and the extended questions $\mathcal{D}$@K generated at the K-th iteration. 
The severity of data contamination ($\Delta$) is measured by the improvement in performance from the fine-tuned model $\mathcal{M}$-sft to the base model $\mathcal{M}$, which is formalized as $\Delta=ACC_{\mathcal{M}\text{-sft}}-ACC_{\mathcal{M}}$.
To mitigate the effects of the LLM's randomness, each question was answered by the LLM 5 times. 
If the LLM produced a correct answer in three or more of these trials, it was considered to have answered the question correctly.
We conduct our experiments on an Ubuntu machine with one 40GB NVIDIA A100 GPU. The results are shown in Table~\ref{tb: exp_result_data_contamination}.

The experiment results indicate that, across various LLMs and benchmarks, fine-tuning with evaluation data leads to a notable improvement in model performance on base questions ($\Delta$-base), particularly evident on the MultiHop-RAG benchmark. 
These findings underscore the risks of data contamination: even when the original model exhibits limited performance on benchmarks, exposure to the benchmark evaluation data can artificially inflate its performance. 
Consequently, the model’s genuine capabilities may be obscured by overestimated benchmark performance, leading to misapplication in scenarios beyond the actual capabilities.

In contrast, when evaluated on extended questions generated by JudgeAgent, the fine-tuned models and base models show little difference in performance.
Additionally, fine-tuning even resulted in a decline on the MedQA benchmark. 
These results suggest that under the JudgeAgent dynamic evaluation paradigm, the generated questions maintain the validity of the evaluation, even when the original questions have been exposed to the target model. 
Furthermore, under the same setting of LLM and benchmark, there is a marked gap between the performance gain on base questions and extended questions after fine-tuning, demonstrating that JudgeAgent exhibits considerable resilience to data contamination.

\begin{table}[ht]
\begin{center}
\begin{tabular}{l|ccc}
\toprule
\textbf{Target} & \textbf{ACC1} & \textbf{ACC2}$_d$ & \textbf{ACC2}$_f$ \\
\midrule
Qwen3 & 91.71 & 99.18 & 96.38 \\
GLM4-Flash & 80.09 & 98.37 & 92.82 \\
GPT-4.1 & 84.97 & 98.28 & 92.44 \\
\bottomrule
\end{tabular}
\end{center}
\caption{The comparative results on MedQA of directly providing correct answers (ACC2$_d$) and providing feedback from JudgeAgent (ACC2$_f$). ACC1 represents the accuracy of target models before evaluation. All values are percentages. }
\label{tb: exp_compare_direct_and_feedback}
\vspace{-1em}
\end{table}

Additionally, we conducted a comparative experiment providing the target models with both standard evaluation feedback from JudgeAgent and evaluation feedback containing the correct answers as cheating information, investigating the risk of data contamination during the evaluation. The results are presented in Table~\ref{tb: exp_compare_direct_and_feedback}. As can be observed, when the correct answers are directly provided, nearly all LLMs achieve almost perfect accuracy. This further illustrates the risk of data contamination in existing static benchmarks. It also demonstrates that the evaluation feedback from JudgeAgent does not contain cheating information tailored specifically to the seed questions

\subsection{Quality of Evaluation Questions}
\label{appendix: question_quality}

We use Dingo\footnote{\url{https://github.com/MigoXLab/dingo}} to check the factuality of evaluation questions generated by JudgeAgent in multi-turn interactions. 
The results are shown in Table~\ref{tab: fact_check}, $N_{eq}$ represents the number of generated questions in Staget Interactive Extension.

\begin{table}[h]
    \centering
    \footnotesize
    \setlength{\tabcolsep}{3pt}
    \resizebox{\linewidth}{!}{
        \begin{tabular}{l|c|c|cc}
            \toprule
            \multirow{2}{*}{\textbf{Datasets}} & \multirow{2}{*}{$N_{eq}$} & \multirow{2}{*}{avg. fact ratio} & \multicolumn{2}{c}{fact check passed}\\
             & & & $N_{passed}$ & passed ratio \\
            \midrule
            MedQA & 3819 & 97.93\% & 3680 & 96.36\% \\
            MultiHop-RAG & 7668 & 94.38\% & 7185 & 93.70\% \\
            \bottomrule
        \end{tabular}
    }
    \caption{The fact check results by Dingo.}
    \label{tab: fact_check}
    \vspace{-5pt}
\end{table}

As shown in Table~\ref{tab: fact_check}, in Dingo’s fact-checking task, the extended questions from JudgeAgent maintain a high fact ratio and detection pass ratio (using Dingo’s default fact ratio threshold of 80\%). 
The average fact ratio is above 94\%, while the pass ratio exceeds 93\% for both MedQA and MultiHop-RAG. 
The results demonstrate that the most generated questions exhibit strong factuality, ensuring that the dynamic evaluation results are based on high-quality questions, and the probability of an evaluated LLM being incorrectly judged as wrong when its answer is correct remains low. 

As for the diversity of evaluation questions, the experiment results of data contamination in Table~\ref{tb: exp_result_data_contamination} in Appendix~\ref{appendix: exp_data_contamination} can indirectly demonstrate that contaminated models perform poorly on synthesized questions, which indirectly confirms that synthesized questions differ from seed questions and that their diversity is assured.

\subsection{Cross Validation of the Evaluation Suggestions}
\label{appendix: cross_validation}
\textbf{Do the suggestions provided by JudgeAgent only take effect for the seed questions?} 
To address this question, we designed cross-validation experiments from two perspectives.

\begin{figure*}
    \centering
    \includegraphics[width=\linewidth]{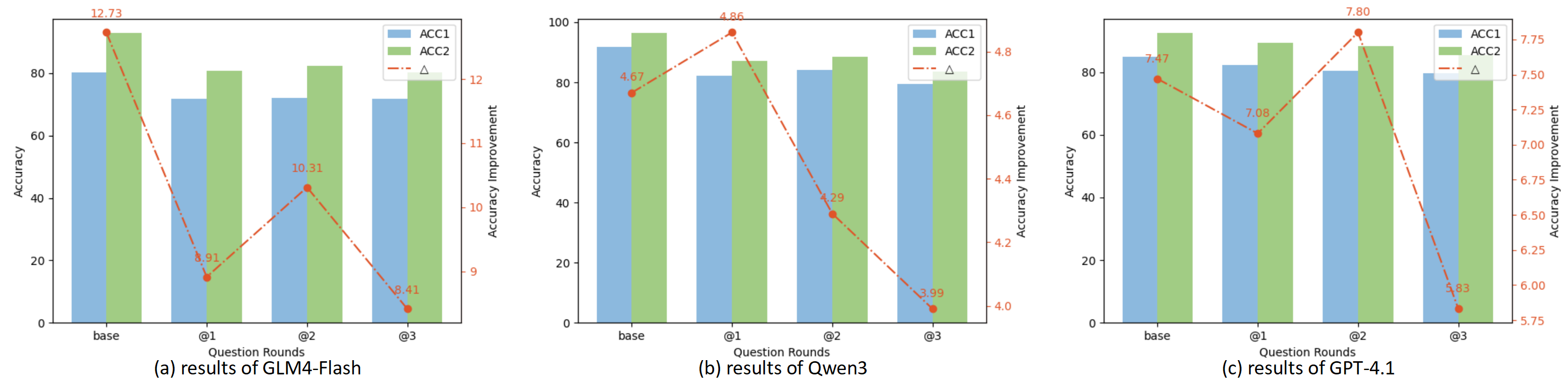}
    \caption{The cross-validation results of extended questions versus seed questions on MedQA. ACC1 and ACC2 indicate the accuracy before and after evaluation. $\Delta=\text{ACC2}-\text{ACC1}$ refers to the overall accuracy improvement. All the values are percentages. @K represents the results of the questions expanded at the K-th round.}
    \label{fig: cross-validation-follow-up}
    \vspace{-1em}
\end{figure*}

First, considering that the evaluation suggestions are derived from a comprehensive evaluation of the responses to both the seed questions and their extended questions, we evaluated and compared the accuracy improvement of the suggestions on the extended questions versus the seed questions, aiming at verifying that the suggestions do not contain "cheating information" specific to seed questions in the scenario after evaluation. 
The results are shown in Figure~\ref{fig: cross-validation-follow-up}.

Secondly, we transferred the evaluation suggestions derived from seed questions to other questions with related knowledge concepts in the benchmark, to verify the effectiveness of the suggestions within the same knowledge domain. 
Specifically, we categorized questions based on the knowledge entities they contain, formalized as follows: 
given the context graph $\mathcal{G}$, for two questions $q_1$ and $q_2$ with knowledge entities $E_1=\{e|e\in q_1\land e\in\mathcal{G}\}$ and $E_1=\{e|e\in q_2\land e\in\mathcal{G}\}$, if $E_1$ and $E_2$ have a non-empty intersection, then $q_1$ and $q_2$ are considered related questions. 
In this experiment, the suggestions for a question were constructed from the suggestions of its related questions, excluding suggestions from the question itself, to assess and validate the transferability of the suggestions provided by the JudgeAgent. We screened out questions without relevant questions and those with only relevant questions based on non-knowledge entities, such as male, female, 2 years, etc. The results are shown in Table~\ref{tb: exp_result_cross_validation}.

\begin{table*}[ht]
\begin{center}
\begin{tabular}{l|c|cccc|cccc}
\toprule
\multirow{2}{*}{Target} & \multirow{2}{*}{ACC1} & \multicolumn{4}{c|}{Non-transfer} & \multicolumn{4}{c}{Transfer} \\ 
 & & ACC2 & CR$\uparrow$ & CtE$\downarrow$ & $\Delta\uparrow$ & ACC2 & CR$\uparrow$ & CtE$\downarrow$ & $\Delta\uparrow$  \\
\midrule
GLM4-Flash & 78.42 & 87.49 & 18.30 & 9.23 & 9.07 & 85.56 & 16.98 & 9.84 & 7.13 \\
GPT-4.1 & 84.21 & 91.09 & 10.23 & 3.35 & 6.88 & 89.78 & 9.50 & 3.93 & 5.57 \\
Qwen3 & 92.44 & 96.06 & 5.96 & 2.34 & 3.62 & 95.17 & 5.17 & 2.44 & 2.73\\
\bottomrule
\end{tabular}
\end{center}
\caption{The cross-validation results of transferring suggestions to related questions. All the values are percentages. \textit{Non-transfer} refers to suggestions being applied to the seed questions, whereas \textit{Transfer} refers to them being applied to related questions.}
\label{tb: exp_result_cross_validation}
\end{table*}

First, we analyze the difference in the effectiveness of evaluation suggestions on seed questions versus extended questions. 
As shown in Figure~\ref{fig: cross-validation-follow-up}, suggestions effectively improve performance for both seed questions and extended questions, with a relatively small gap in the degree of improvement. 
Notably, when Qwen3 and GPT-4.1 are used as target models, the first round of Qwen3 and the second round of GPT-4.1 exhibit even greater improvement than seed questions. 
These results indicate that although the suggestions only supplement knowledge for several key concepts, such as the case in Figure~\ref{fig: case_detail}, such concise suggestions can still benefit both seed and extended questions, demonstrating that JudgeAgent is capable of identifying and addressing knowledge gaps in the target model, rather than simply providing "cheating information" specific to seed questions.

Furthermore, it is observed that the effectiveness of suggestions for third-round questions is consistently low in the experiments. 
This may be because, by the third round, knowledge path sampling has expanded beyond the scope of knowledge related to seed questions to a broader range. 
As a result, the generated questions diverge more significantly in core knowledge from earlier questions, thereby reducing the effectiveness of the knowledge guidance provided in the suggestions.

Next, we analyze the difference in the effectiveness of suggestions on questions that share the same knowledge concepts. 
As shown in Table~\ref{tb: exp_result_cross_validation}, compared to their effectiveness on seed questions, the suggestions exhibit a slight decrease when applied to related questions, as indicated by a decline in CR and $\Delta$, and an increase in CtE. 
But the difference is minor, and the improvement remains notable, suggesting that the suggestions can be effectively transferred to other questions involving the same knowledge concepts. 
This further demonstrates that JudgeAgent can provide suggestions that do not simply serve as “cheating information” specific to seed questions.

However, the slight decline in evaluation effectiveness also indicates that the knowledge guidance provided in the suggestions is not fully aligned with the related questions. 
This may be because the overlapping entities between these related questions and seed questions do not correspond to core knowledge concepts. 
For example, suggestions centered on "blood type" may be transferred to a question where "serum" is the core knowledge concept, resulting in a partial mismatch.

The above experiment results demonstrate that the evaluation suggestions provided by JudgeAgent are not only applicable to seed questions but can also be transferred to other questions that share relevant core knowledge concepts.

\subsection{Supplementary Parameter Analysis}
\label{appendix: batch_size_analysis}
\begin{figure*}[htbp]
    \centering
    \includegraphics[width=\linewidth]{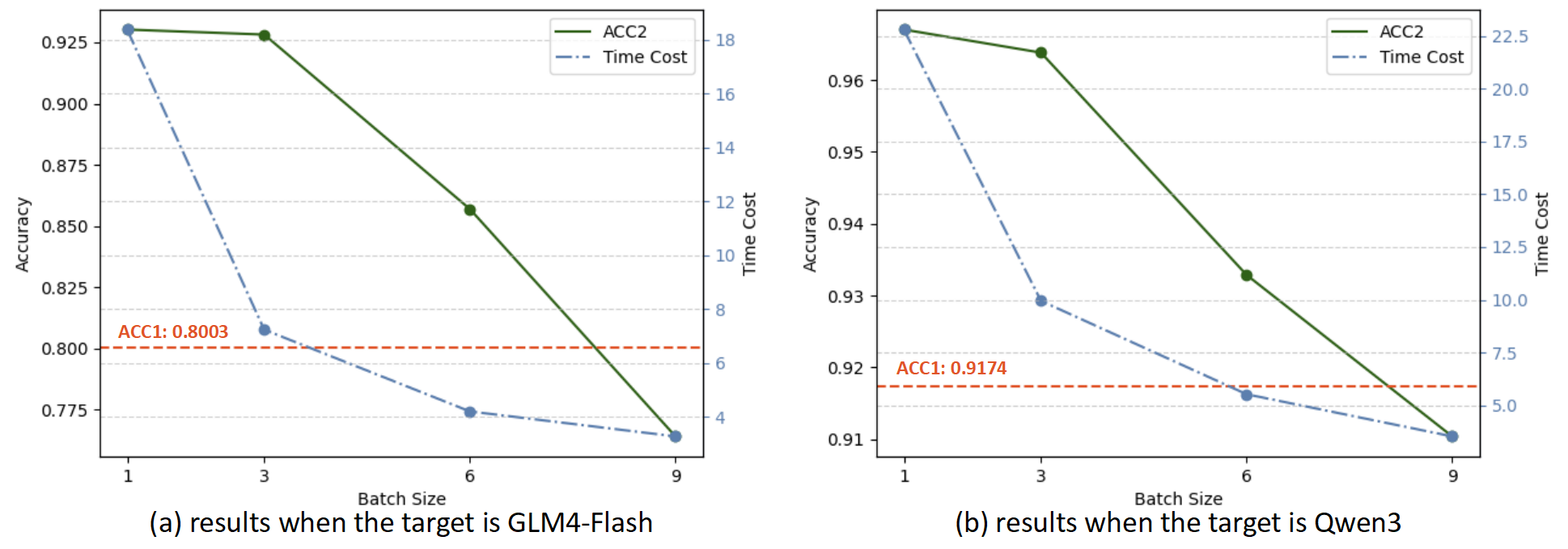}
    \caption{The results of different batch sizes on MedQA. ACC1 and ACC2 indicate the accuracy before and after receiving evaluations. Time Cost is the average time consumption for each question.}
    \label{fig: batch_size_analysis}
\end{figure*}

\textbf{What is the impact of batch size in the \textit{Benchmark Grading} stage on the evaluations of the JudgeAgent?}
In \textit{Benchmark Grading} stage, questions are divided into batches to comprehensively assess the target's capabilities in a base level. 
These batches are also the basic units for question extension and evaluation feedback.
Given a fixed rounds, the batch size is inversely proportional to the number of batches, extended questions, and evaluation suggestions, thereby influencing the time and resource consumption of the entire evaluation process.
\textbf{Can the batch size be maximized to reduce resource consumption while maintaining the effectiveness of evaluations?} To address this question, we conducted a parameter analysis experiment, examining the evaluation effectiveness and time consumption under different batch sizes. The results are shown in Figure~\ref{fig: batch_size_analysis}.

As observed from the trend of the curves in Figure~\ref{fig: batch_size_analysis}, both the target's accuracy after receiving evaluation suggestions (ACC2) and the average time consumption per question for evaluation (Time Cost) decrease as the batch size increases. 
Among them, the decline rate in ACC2 gradually accelerates with larger batch sizes, while the decline rate in Time Cost gradually slows, exhibiting a marginal effect. 
The reason for the marginal effect in Time Cost lies in the fact that the time required for the target model to answer the seed questions, which is a component of the overall evaluation process, varies little with changes in batch size. 
As a result, there is a threshold beyond which further reductions in time cost have diminishing returns.

The decline in ACC2 with increasing batch size can be attributed to the expansion of the question's knowledge domain. 
As the batch size grows, the generated questions during evaluation become more heterogeneous and less coherent with the knowledge relevant to seed questions, making it difficult for JudgeAgent to identify appropriate knowledge guidance from the dispersed question-answer pairs. 
Moreover, when the batch size exceeds the number of extension rounds, the disorder of knowledge scopes intensifies more rapidly, ultimately leading to a sharp drop in accuracy. 
The results in Figure~\ref{fig: batch_size_analysis} show that when the batch size reaches 9, the evaluation suggestions even become counterproductive (ACC2 $<$ ACC1), interfering with the normal reasoning of the target model.

Therefore, considering both the evaluation time cost and effectiveness, we selected a batch size of 3 in our experiments as a balanced choice.

\section{Algorithm}
\label{appendix: algorithm}

To clarify the entire workflow of JudgeAgent, we use pseudocode to show the dynamic evaluation process in Algorithm~\ref{pcode: judgeagent}, the validation process of evaluation in Algorithm~\ref{pcode: validation}, the construction of context graph in Algorithm~\ref{pcode: context_graph}, and the generation of extended questions in Algorithm~\ref{pcode: extension}.

\begin{algorithm*}[t]
\caption{Dynamic evaluation process of JudgeAgent}
\label{pcode: judgeagent}
\begin{algorithmic}[1]
\State \textbf{Input:} Target LLM $\mathcal{M}_t$, Base dataset $\mathcal{D}=\{(q_i,a_i,d_i)\}_{i=1}^N$ (each item include question $q$, answer $a$, and difficulty $d$), Knowledge bases $\mathcal{K}=\{k_1,k_2,...\}$, Core LLM of JudgeAgent $\mathcal{M}_c$, Predefined batch capacity $N_B$, Max extension round $RND_{e}$, Max hop of sampling $H$
\State \textbf{Output:} Evaluation Score $sc$, Batched dataset with suggestions $\mathcal{D}_S=\{(\mathcal{B}_1\,,\,s_1)\,,\,\dots\}$
\Statex \textit{// Construct context graph}
\State $\mathcal{G}\gets$\Call{Construct\_Context\_Graph}{$\mathcal{K}\,,\,\mathcal{M}_c$}
\Statex \textit{// Split base dataset into batches}
\State $\{\mathcal{B}_1=\{(q_{1i},a_{1i},d_{1i})\}_{i=1}^{N_B}\,,\dots\}\gets$\Call{Split\_Batches}{$\mathcal{D}\,,\,N_B$}
\Statex \textit{// Begin Evaluation}
\State $sc\gets0\ ,\ N_{total}\gets0$
\State $\mathcal{D}_S\gets\{\}$
\For{$\mathcal{B}\gets\{\mathcal{B}_1\,,\,\mathcal{B}_2\,,\,\dots\}$}
    \State $sc_\mathcal{B}\gets0$
    \State $RND_{total}\gets0$
    \State $\mathcal{Q}_{tested}\gets\{\}$
    \Statex \textit{// Stage1: Benchmark Grading}
    \For{$(q,a,d)\gets\mathcal{B}$}
        \State Get answer $a_\mathcal{M}\gets$\Call{Query\_LLM}{$q\,,\,\mathcal{M}_t$}
        \If{$a_\mathcal{M}$ is correct based on $q$ and $a$}
            \State $sc_\mathcal{B}\gets sc_\mathcal{B}+$\Call{Difficulty\_Score}{$d$}
        \EndIf
        \State $RND_{total}\gets RND_{total}+1$
        \State Add $(q,a,d,a_\mathcal{M})$ to $\mathcal{Q}_{tested}$
    \EndFor
    \State Decide difficulty $d_e\gets$\Call{Decide\_Difficulty}{$sc_\mathcal{B}\,,\,RND_{total}$}
    \Statex \textit{// Stage2: Interactive Extension}
    \For{$i\gets\{1,2,\dots,RND_e\}$}
        \State $(q_e,a_e,t)\gets$\Call{Generate\_Extended\_Questions}{$\mathcal{B}\,,\,\mathcal{G}\,,\,\mathcal{M}_c\,,\,H\,,\,d_e$}
        \State Get answer $a_\mathcal{M}\gets$\Call{Query\_LLM}{$q_e\,,\,\mathcal{M}_t$}
        \If{$a_\mathcal{M}$ is correct based on $q_e$ and $a_e$}
            \State $sc_\mathcal{B}\gets sc_\mathcal{B}+$\Call{Difficulty\_Score}{$d_e$}
        \EndIf
        \State $RND_{total}\gets RND_{total}+1$
        \State Add $(q_e,a_e,d_e,a_\mathcal{M})$ to $\mathcal{Q}_{tested}$
        \State Decide difficulty $d_e\gets$\Call{Decide\_Difficulty}{$sc_\mathcal{B}\,,\,RND_{total}$}
    \EndFor
    \Statex \textit{// Stage3: Evaluation Feedback}
    \State Get evaluation suggestions $s\gets$\Call{Evaluate}{$\mathcal{Q}_{tested}\,,\,\mathcal{M}_c$}
    \State Add $(\mathcal{B},s)$ to $\mathcal{D_S}$
    \State $sc\gets sc\,+\,sc_\mathcal{B}$
    \State $N_{total}\gets N_{total}+RND_{total}$
\EndFor
\State $sc\gets sc\,/\,N_{total}$
\State \Return $sc\ ,\ \mathcal{D}_S$
\end{algorithmic}
\end{algorithm*}

\begin{algorithm*}[t]
\caption{Validation process of evaluation results from JudgeAgent}
\label{pcode: validation}
\begin{algorithmic}[1]
\State \textbf{Input:} Target LLM $\mathcal{M}_t$, Batched dataset with suggestions $\mathcal{D}_S=\{(\mathcal{B}_1\,,\,s_1)\,,\,\dots\}$, in which $\mathcal{B}_i=\{(q_{i1},a_{i1},d_{i1})\,,\,\dots\}$ is a batch of base dataset $\mathcal{D}$, and $\mathcal{S}_i=\{s_{i1}\,,\,\dots\}$ is relevant suggestions from JudgeAgent.
\State \textbf{Output:}Accuracy of target LLM before evaluation $acc_1$, Accuracy after evaluations $acc_2$, Correction Rate $cr$, Correct-to-Error Rate $ce$
\Statex \textit{// Initialize counter of questions}
\State $N_{acc1}\gets0\ ,\ N_{acc2}\gets0\ ,\ N_{cr}\gets0\ ,\ N_{ce}\gets0\ ,\ N_{total}\gets0$
\For{$(\mathcal{B},s)\gets\mathcal{D}_S$}
    \State $N_\mathcal{B}\gets$\Call{len}{$\mathcal{B}$}
    \State $N_{total}\gets N_{total}+N_\mathcal{B}$
    \For{$i\gets\{1,2,\dots,N_\mathcal{B}\}$}
        \State $(q,a,d)\gets\mathcal{B}[i]$
        \State Get answer1 $a_1\gets$\Call{Query\_LLM}{$q\,,\,\mathcal{M}_t$}
        \State Get answer2 $a_2\gets$\Call{Query\_LLM\_with\_Suggestions}{$q\,,\,s\,,\,\mathcal{M}_t$}
        \State correct1$\gets$whether $a_1$ is correct based on $q$ and $a$
        \State correct2$\gets$whether $a_2$ is correct based on $q$ and $a$
        \If{correct1}
            \State $N_{acc1}\gets N_{acc1}+1$
            \If{not correct2}
                \State $N_{ce}\gets N_{ce}+1$
            \EndIf
        \EndIf
        \If{correct2}
            State $N_{acc2}\gets N_{acc2}+1$
            \If{not correct1}
                \State $N_{cr}\gets N_{cr}+1$
            \EndIf
        \EndIf
    \EndFor
\EndFor
\State $acc_1\gets N_{acc1}/N_{total}\ ,\ acc_2\gets N_{acc2}/N_{total}\ ,\ cr\gets N_{cr}/N_{total}\ ,\ ce\gets N_{ce}/N_{total}$
\State \Return $acc_1\ ,\ acc_2\ ,\ cr\ ,\ ce$
\end{algorithmic}
\end{algorithm*}

\begin{algorithm*}[t]
\caption{Construction process of context graph}
\label{pcode: context_graph}
\begin{algorithmic}[1]
\State \textbf{Input:} Knowledge Base of Dataset $\mathcal{K}=\{k_1\,,\,k_2\,,\,\dots\}$, an LLM $\mathcal{M}$
\State \textbf{Output:} Context Graph $\mathcal{G}=(\mathcal{N},\mathcal{E})$, in which $\mathcal{N}$ is the node set, and $\mathcal{E}$ is the edge set.
\State $\mathcal{N}\gets\{\}\ ,\ \mathcal{E}\gets\{\}$\qquad\textit{// Initialize node set and edge set as empty dictionary}
\For{$k\gets\mathcal{K}$}
    \Statex \textit{// Spliting text into chunks}
    \State $C=\{c_1,c_2,\dots\}\gets$\Call{Split\_to\_Chunks}{$k$}
    \Statex \textit{// Prompting LLM to label entities}
    \State $P_e\gets\emptyset$
    \For{$c\gets\{c_1,c_2,\dots\}$}
        \State $S_e=\{e_1,e_2,\dots\}\gets$\Call{Label\_Entity}{$c\,,\,\mathcal{M}$}
        \For{$e\gets S_e$}
            \If{$e$ not in $\mathcal{N}$}
                \State $\mathcal{N}[e]\gets(e\,,\,\mathcal{C}=\emptyset\,,\,\mathcal{D}=\emptyset)$
                \State $\mathcal{E}[e]\gets\emptyset$
            \EndIf
            \State Add $c$ to set $\mathcal{N}[e].\mathcal{C}$
            \State Add $k$ to set $\mathcal{N}[e].\mathcal{D}$
            \State Expand set $\mathcal{E}[e]$ with $S_e$
        \EndFor
        \State Expand set $P_e$ with $S_e$
    \EndFor
    \For{$e\gets P_e$}
        \State Expand set $\mathcal{E}[e]$ with $P_e$
    \EndFor
\EndFor
\State $\mathcal{G}\gets(\mathcal{N}\,,\,\mathcal{E})$
\State \Return $\mathcal{G}$
\end{algorithmic}
\end{algorithm*}

\begin{algorithm*}[t]
\caption{Generation process of extended questions}
\label{pcode: extension}
\begin{algorithmic}[1]
\State \textbf{Input:} Base Question $\mathcal{Q}=\{q_1,q_2,\dots\}$, Context Graph $\mathcal{G}=(\mathcal{N},\mathcal{E})$, LLM $\mathcal{M}$, Max hop of path $H$, Difficulty $d$
\State \textbf{Output:} Extended question with its answer and background text $(q_e\,,\,a_e\,,\,t)$
\Statex \textit{// Prompting LLM to label entities from }$\mathcal{Q}$
\State $S_e\gets\{\}$
\For{$q\gets\mathcal{Q}$}
    \State $set_e=\{e_1,e_2,\dots\}\gets$\Call{Label\_Entity}{$q\,,\,\mathcal{M}$}
    \State Add $set_e$ to $S_e$
\EndFor
\State $e\gets$\Call{Random\_Sample}{$\{e_1,e_2,\dots\}\,,\,1$}
\Statex \textit{// Sample knowledge paths}
\State $e'\gets$the most similar entity in $\mathcal{N}$ of $\mathcal{G}$
\State $E_{visited}\gets\{e'\}$
\State $t\gets$\Call{Random\_Sample}{$\mathcal{N}[e'].C\,,\,1$}
\For{$i\gets\{1,2,\dots,H-1\}$}
    \State $E_{candidate}\gets$the most similar 5 entities to $e'$ in $\mathcal{E}[e']$ that not in $E_{visited}$
    \State $e'\gets$\Call{Random\_Sample}{$E_{candidate}\,,\,1$}
    \State $C_{candidate}\gets$the most similar 5 chunks to $q$ in $\mathcal{N}[e'].C$
    \State $c\gets$\Call{Random\_Sample}{$C_{candidate}\,,\,1$}
    \State Concatenate $c$ to new line of $t$
    \State Add $e'$ to $E_{visited}$
\EndFor
\State $(q_e, a_e)\gets$\Call{Generate\_Question}{$t\,,\,\mathcal{M}$}
\State \Return $(q_e\,,\,a_e\,,\,t)$
\end{algorithmic}
\end{algorithm*}
\section{Prompts}
\label{appendix: prompts}

We show the detailed prompt for Entity Extraction in Prompt~\ref{tb: prompt_entity_extraction}, the prompt for generating extended questions in Prompt~\ref{tb: prompt_generation_questions}, the prompt for evaluating the performance of the target LLM in Prompt~\ref{tb: prompt_evaluation}, and the prompt for querying the target LLM with suggestions in Prompt~\ref{tb: prompt_query}.

\begin{table*}[htbp]
    \centering
    \begin{adjustbox}{max width=\textwidth}
    \begin{tabular}{p{\textwidth}}
        \toprule
        {\bfseries Prompt for Entity Extraction}\cr
        \midrule
        Please identify and label the entities in the following multiple sentences, and return the entity labeling results for each sentence.\cr
        The results for each sentence should be independent, in JSON format, containing the sentence number, sentence text, and the list of recognized entities (including entity text, type, and position).\cr
        Return format is a dictionary, with only one key 'labeled\_data', and the value is a list, each element is a dictionary containing the sentence text and the entity list.\cr
        \{\{\cr
        \qquad"labeled\_data":\cr             
        \qquad[\cr
            \qquad\qquad\{\{"text":"Sentence 1", "entity\_list": [\{\{"entity\_text": "", "entity\_type": ""\}\}]\}\}, \cr
            \qquad\qquad\{\{"text":"Sentence 2", "entity\_list": [\{\{"entity\_text": "", "entity\_type": ""\}\}]\}\}, \cr
            \qquad\qquad...\cr
        \qquad]\cr
        \}\}\cr
        Notice that "text" should be only the sentence, not the whole article.
        Sentence list:\cr
        \{ \textit{Sentences} \}\cr
        \bottomrule
    \end{tabular}
    \end{adjustbox}
    \caption{Prompt for entity extraction.}
    \label{tb: prompt_entity_extraction}
\end{table*}

\begin{table*}[htbp]
    \centering
    \begin{adjustbox}{max width=\textwidth}
    \begin{tabular}{p{\textwidth}}
        \toprule
        {\bfseries Prompt for Generating Extended Questions}\cr
        \midrule
        As an interviewer, you are tasked with designing questions based on the provided texts. Your role involves crafting questions and correct answers that fulfill the following criteria:\cr\cr
        1. **Focus on the Entity**: Ensure all questions consistently center around the specified entity from the article.\cr
        2. **Ensure Accuracy and Conciseness of Answers**: Verify that the provided answer is both correct for your designed question within the context and logic of the given text fragments, and ensure the answer is sufficiently concise—presented as a word or phrase, avoiding redundancy.\cr
        3. **Conform to difficulty requirements**: You need to design questions for the required difficulty levels, with specific requirements as follows:\cr
        \qquad(1). [easy]: **Encourage Knowledge Memorization**, design questions that assess whether respondents have memorized relevant knowledge. Create questions by directly extracting and blanking out content from the given passage.\cr
        \qquad(2). [medium]: **Encourage Knowledge Comprehension**: Design questions that prompt respondents to dissect and comprehend concepts involved in the topic. Avoid assessing only superficial knowledge retention.\cr
        \qquad(3). [hard]: **Encourage Knowledge Deep Analysis**: Design questions that prompt respondents to engage in deep thinking and analysis. Avoid merely testing knowledge recall or conceptual comprehension; do not simply extract fragments from the given passage to create fill-in-the-blank items. Encourage respondents to focus on entities within the question and employ logical skills for complex reasoning.\cr\cr
        Here are examples:\cr
        \{ \textit{Examples} \}\cr\cr
        Now, given the following text fragments:\cr
        \{ \textit{context} \}\cr\cr
        Based on the provided texts, please generate questions by following the requirements above and referencing the examples.\cr
        Output in the specified JSON format below:
        \{\{\cr
        \qquad"generated\_question":\cr             
        \qquad[\cr
            \qquad\qquad\{\{\cr
            \qquad\qquad\qquad"question": "Generated Question",\cr
            \qquad\qquad\qquad"answer": "Correct Answer of Generated Question"\cr
            \qquad\qquad\}\},\cr
            \qquad...\cr
        \qquad]\cr
        \}\}\cr
        \bottomrule
    \end{tabular}
    \end{adjustbox}
    \caption{Prompt for generating extended questions based on different difficulties.}
    \label{tb: prompt_generation_questions}
\end{table*}

\begin{table*}[htbp]
    \centering
    \begin{adjustbox}{max width=\textwidth}
    \begin{tabular}{p{\textwidth}}
        \toprule
        {\bfseries Prompt for Evaluation}\cr
        \midrule
        The following is the performance of an LLM in answering a series of questions:\cr\cr
        \{ \textit{list of questions, correct answers, and LLM's answers} \}\cr\cr
        
        Please evaluate and analyze the interviewee's performance based on the above performance using concise language from the following perspectives, and provide suggestions that help the LLM answer the same questions better. Suggestions should provide specific and detailed guidance on logical thinking steps, required knowledge, and abilities, ensuring the LLM can answer correctly for the same questions.\cr
        Output in the following JSON format:\cr
        \{\{\cr
        \qquad"flaws\_knowledge": "The lack of background knowledge.",\cr
        \qquad"flaws\_capability": "The flaws in logic and capability.",\cr
        \qquad"comprehensive\_performance": "The Comprehensive performance of all questions.",\cr
        \qquad"suggestions": "Suggestions that help the LLM answer questions bette"\cr
        \}\}\cr
        \bottomrule
    \end{tabular}
    \end{adjustbox}
    \caption{Prompt for evaluating the performance of the target LLM.}
    \label{tb: prompt_evaluation}
\end{table*}

\begin{table*}[htbp]
    \centering
    \begin{adjustbox}{max width=\textwidth}
    \begin{tabular}{p{\textwidth}}
        \toprule
        {\bfseries Prompt for Querying LLM with suggestions}\cr
        \midrule
        Please complete the following question:\cr
        [question]: \{ \textit{question} \}\cr\cr
        
        In your previous responses to these questions, the interviewer has provided the following suggestions for you to help you answer better: \cr
        [suggestions]: \{ \textit{suggestions} \}\cr\cr
        
        Please consider the above [suggestions], and answer the above [question], in the following JSON format:\cr
        \{\{"answer": "Your answer"\}\}\cr
        \bottomrule
    \end{tabular}
    \end{adjustbox}
    \caption{Prompt for querying the target LLM with suggestions.}
    \label{tb: prompt_query}
\end{table*}
\section{Detailed Content of Case Study}
\label{appendix: case_study}

\begin{figure*}
    \centering
    \includegraphics[width=\linewidth]{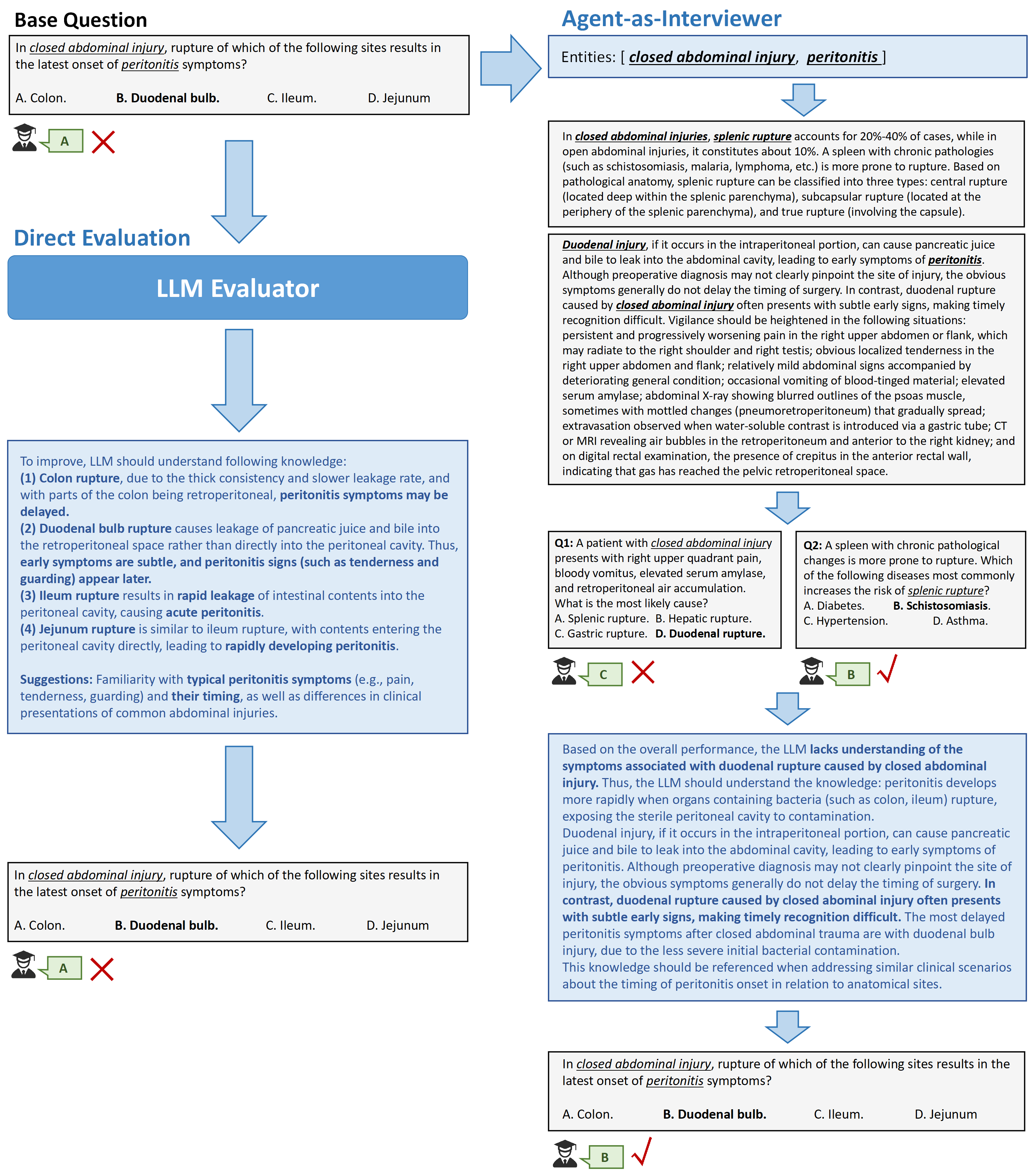}
    \caption{The detailed content of the case.}
    \label{fig: case_detail}
\end{figure*}

The detailed content of the comparative case in section~\ref{subsec: 5_6_parameter} is shown in Figure~\ref{fig: case_detail}.

\end{document}